\documentclass{article}

\usepackage{arxiv}

\usepackage[utf8]{inputenc} 
\usepackage[T1]{fontenc}    
\usepackage{hyperref}       
\usepackage{url}            
\usepackage{booktabs}       
\usepackage{amsfonts}       
\usepackage{nicefrac}       
\usepackage{microtype}      
\usepackage{lipsum}		
\usepackage{graphicx}
\usepackage{natbib}
\usepackage{doi}
\usepackage{blindtext}
\usepackage[T1]{fontenc}
\usepackage[utf8]{inputenc}
\usepackage{babel}
\usepackage[font=small,labelfont=bf]{caption}
\usepackage{graphicx}
\usepackage{subcaption}

\title{Development of Interpretable Machine Learning Models to Detect Arrhythmia based on ECG Data}

\author{\href{https://www.researchgate.net/profile/Shourya-Verma-2}{Shourya Verma} \\
	Universität Heidelberg\\
	\texttt{shourya.verma@stud.uni-heidelberg.de} \\
}

\renewcommand{\shorttitle}{Interpretable ML to Diagnose Arrhythmia}

\hypersetup{
pdftitle={Development of Interpretable Machine Learning Models to Detect Arrhythmia based on ECG Data},
pdfauthor={Shourya Verma},
pdfkeywords={Machine Learning, Interpretability, Arrhythmia, Grad-CAM},
}

\begin{document}
\maketitle

\begin{abstract}
The analysis of electrocardiogram (ECG) signals can be time consuming as it is performed
manually by cardiologists. Therefore, automation through machine learning (ML) classification
is being increasingly proposed which would allow ML models to learn the features of a heartbeat
and detect abnormalities. The lack of interpretability hinders the application of Deep Learning
in healthcare. Through interpretability of these models, we would understand how a machine
learning algorithm makes its decisions and what patterns are being followed for classification. This thesis builds Convolutional Neural Network (CNN) and Long Short-Term Memory (LSTM) classifiers based on state-of-the-art models and compares their performance and
interpretability to shallow classifiers. Here, both global and local interpretability methods are
exploited to understand the interaction between dependent and independent variables across the
entire dataset and to examine model decisions in each sample, respectively. Partial Dependence
Plots, Shapley Additive Explanations, Permutation Feature Importance, and Gradient Weighted
Class Activation Maps (Grad-Cam) are the four interpretability techniques implemented on
time-series ML models classifying ECG rhythms. In particular, we exploit Grad-Cam, which is a
local interpretability technique and examine whether its interpretability varies between correctly
and incorrectly classified ECG beats within each class. Furthermore, the classifiers are evaluated
using K-Fold cross-validation and Leave Groups Out techniques, and we use non-parametric
statistical testing to examine whether differences are significant. It was found that Grad-CAM was the most effective interpretability technique at explaining predictions of proposed CNN and LSTM models. We concluded that all high performing classifiers looked at the QRS complex of the ECG rhythm when making predictions.
\end{abstract}


\section{Introduction}

The Electrocardiogram (ECG) is a medical test that detects cardiac abnormalities by measuring
the electrical signals generated by the heart during contraction. This test is the most accessible
and inexpensive tool for diagnosing conditions like arrhythmia. Arrhythmia is a cardiac abnormality related to the rate and rhythm of the heartbeat. Despite being the most frequently
used diagnosing tool, the rates of ECGs misdiagnosis are still too high. A study from 2016
showed that approximately one in three patients out of 550,000 had their ECGs misdiagnosed
and misinterpreted \cite{RN54}. This study provides an insight that ECG misrepresentations
may have far reaching health diagnostic ramifications. Hence, interpretable ECG algorithms
need to be more accurate and explainable so misdiagnoses can be swiftly caught and removed
before affecting a patient.

Traditionally, the analysis of these signals can take time as it is performed by cardiologists. Therefore, automation through Machine Learning (ML) classification is being increasingly proposed
which would allow ML models to learn the features of a heartbeat and detect abnormalities.
The models however fail to recognize unseen ECG data from new datasets while also facing
interpretability and explanation problems. These gaps in interpretability of models occur due to
the ’black-box’ nature of ML models which render these techniques untrustworthy by clinicians
and in-turn limit their application in patient driven healthcare industry.

This paper investigates these concerns by using and building ML models for ECG classification,
and implementing Global and Local Interpretability techniques on the models. These models
were trained on the MIT-BIH arrhythmia dataset \cite{RN41}, where the ECG
data was split into single beats and each beat was classified into one of eight beat classes. The ML model algorithms included naive Bayes’ and ensemble methods, support vector machines, deep
and recurrent neural networks. The results for all the models are compared using non-parametric
statistical hypothesis testing, and issues with different interpretability techniques are addressed.

Since most interpretability libraries do not support time-series data very well due to its abundance of data points per instance, the ECG beats were sliced into 11 segments where different group of segments represented the morphology of the ECG beats. These explanations provide important insight on how a model comes to make its decision. This is done by analyzing individual classes and examining correct and mis-classified beats within classes. Global interpretability methods like Partial Dependency Plots \cite{RN40}, Shapley Additive Explanations \cite{RN37}, and Permutation Feature Importance \cite{RN25} were implemented, and Local
interpretability approach of Gradient Weighted Class Activation Maps \cite{RN47} was
applied on Convolutional Neural Network and Long Short-Term Memory Network.

The github repository can be found here: \url{https://github.com/shouryaverma/interpretable_ml_ecg}. It contains the source code, the results, and the original thesis document.

\section{Background}

\subsection{ECG Structure and MIT-BIH Dataset}

An ECG test consist of collecting data through the electrical activity of the human cardiovascular
system from various different angles by placing electrodes at different points on the skin. This
non-invasive method of analysing the heartbeat consist of three key features which represent
distinct stages of the heartbeat, i.e the P-wave, which reflects the depolarization of the atria;
the QRS complex, which shows the depolarization of the ventricles; and the T-wave, which
represents the re-polarization of the ventricles. This allows us to detect abnormalities by equating each phase to the normal cardiac cycle. Figure \ref{tab:1} shows the ECG signal representation of a normal beat. These ECG signals are extremely susceptible to high and low frequency noise
which usually occur from baseline wander, misplaced electrode contact, motion artifacts, or
power line interference \cite{RN51}.

\begin{minipage}{\textwidth}
  \begin{minipage}[b]{0.39\textwidth}
    \centering
    \includegraphics[width=1\textwidth]{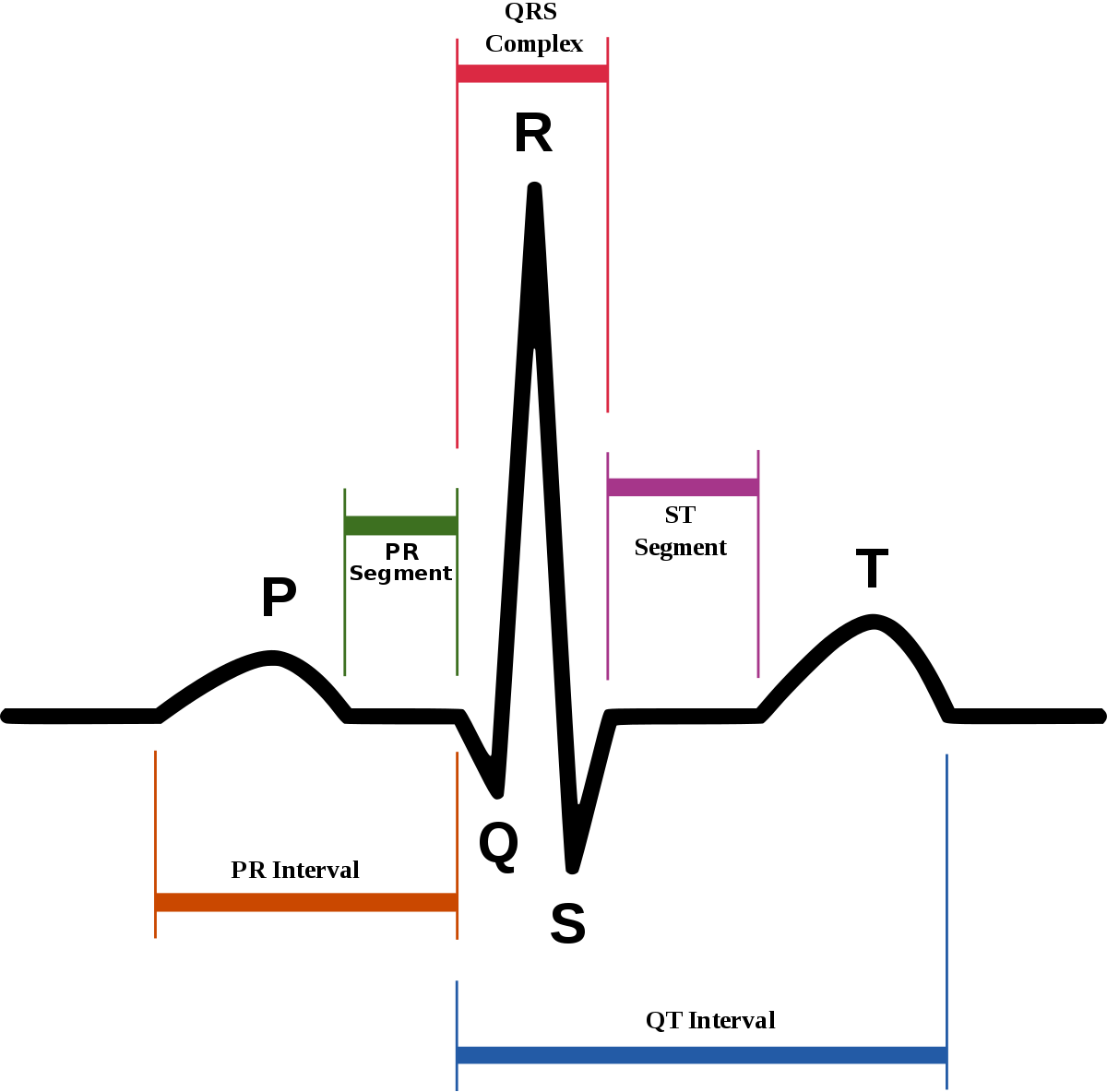}
    \captionof{figure}{Normal ECG Signal}
  \end{minipage}
  \hfill
  \begin{minipage}[b]{0.59\textwidth}
    \centering
    \begin{center}
    \begin{tabular}{ccl}
    \hline \textbf{Class} & \textbf{ID} & \textbf{Beat} \textbf{Description} \\
    \hline $\mathrm{N}$ & 1 & Normal \\
    $\mathrm{L}$ & 2 & Left Bundle Branch Block \\
    $\mathrm{R}$ & 3 & Right Bundle Branch Block \\
    $\mathrm{V}$ & 4 & Premature Ventricular Contraction \\
    $\mathrm{A}$ & 5 & Atrial Premature \\
    $\mathrm{F}$ & 6 & Fusion of Ventricular and Normal \\
    $\mathrm{f}$ & 7 & Fusion of Paced and Normal \\
    $/$ & 8 & Paced \\
    \hline
    \end{tabular}
    \end{center}
    \captionof{table}{Beat Classes, ID Number, and Description}
    \label{tab:1}
    \end{minipage}
  \end{minipage}

The MIT-BIH dataset used for this investigation is a public database consisting of a large number
of beats, and is frequently used for time-series classification research. The MIT-BIH Arrhythmia
Database contains sections of ambulatory ECG recordings, from 47 subjects, digitized at 360
samples per second per channel with 11-bit resolution at 10-mV range on two channels, studied
by the BIH Laboratory. Here 23 recordings were picked at random from a set of 4000 24-hour
ECG recordings collected from a population 60\% of inpatients and 40\% outpatients \cite{RN41}. This data has been pre-annotated and labelled by cardiologists. These different
annotations refer to various normal and abnormal ECG signals which represent different types of
arrhythmia. The dataset consists of ECG signals of various classes, but the eight classes used for
this investigation are ’N’, ’L’, ’R’, ’V’, ’A’, ’F’, ’f’, ’/’. Table \ref{tab:1} shows the description and numerical identification values assigned to these classes.

\subsection{Time-series Classification Based on Machine Learning}

Machine Learning (ML) has increasingly been used in various areas of research, but despite
encouraging results on big-bio datasets, there is still limited number of applications of these algorithms in the healthcare industry. One of the few challenges of ML in healthcare environments
is that there is a large discrepancy of consistency in data characteristics as the global population of patients is diverse and expanding, allowing the data to widely vary between different groups of medical patients \cite{RN34}. This makes the data extremely imbalanced due to the presence of rare conditions within sick patients and a more likely chance for a person to be healthy and asymptomatic. This paper implemented supervised machine learning technique called multi-class classification where the goal is to correctly identify the class an input corresponds to. A regular, but awkward solution to time-series classification is to consider every time point as a unique feature and directly apply a standard learning algorithm. In this procedure, the algorithm disregards information accommodated in the time order of the data. If the feature order were shuffled, the predictions would not differ. It is also usual to use deep and recurrent learning to classify time-series. Long Short-Term Memory and Convolutional Neural Networks are deemed capable of extracting dynamical characteristics of time-series, hence their success.

\subsection{Machine Learning for ECG Classification}

This section briefly reviews state-of-the-art research and literature on ECG classification using
deep and recurrent neural networks which also include different data resampling techniques and
various model architecture. It provides the justification and the overall motivation of creating
proposed neural network architectures and using other shallow algorithms as comparison for
classification and interpretability.

\cite{RN33} and \cite{RN24}, both proposed deep learning convolutional neural
network (CNN) based classification of ECG signals to improve on conventional ML techniques
for analysis of ECG like boosted decision trees, support vector machines and neural network
multi-layer perceptrons. Both papers used the MIT-BIH arrhythmia dataset and their models
displayed an accuracy of 93.40\% and 94.90\% on noise free ECG classification, respectively. Both
papers however only trained and evaluated results on five annotation classes. Despite having
2 less layers than \cite{RN33}, and having a smaller kernel size, \cite{RN24}
reported 1.5\% better accuracy, likely due to the leaky ReLU activation function (as opposed to
normal ReLU) and additional fully connected layer. \cite{RN46} proposed an 11-layer CNN
to classify five classes of beats in the MIT-BIH arrhythmia dataset, using Synthetic Minority
Oversampling Technique (SMOTE) to handle the dataset imbalance. Their network consisted
of four 1D-convolution and max pooling layers, followed by two fully connected ReLU layers
and a fully connected final softmax layer to classify beats into the five classes. The model was
tested using the hold-out method, by randomly splitting their beats into training and testing
datasets, achieving an accuracy of 98.3\% .

\cite{RN28} proposed the use of an LSTM model to classify beats in the MIT-BIH arrhythmia
dataset using a focal loss function. They used a 64 filter LSTM layer followed by two fully
connected layers to classify beats into one of eight classes and trained their model for 350 epochs.
Their model was tested using the hold-out method, with 90\% of training data and 10\% testing
data and achieved an accuracy of 99.30\%. \cite{RN42} improved on this work by
adding a CNN layer to an LSTM model with sequence-to-sequence networks which consist
of recurrent neural network encoder and decoder. SMOTE was used by this paper as well to
address issues with imbalanced ECG data in the MIT-BIH dataset. Their model consisted of three
1D-convolutional layers with stride 1, kernel size 2x1, and ReLU activation. Each convolutional
layer was followed by a max pooling layer of the same kernel size and stride. The output from
these entered a LSTM unit as input, which consisted of the encoder and decoders before creating
the output which was done by a softmax layer. This model outperformed the previous work,
giving accuracy of 99.92\% on noise free, intra-patient, ECG signal paradigm.

Following on from work discussed in this section, this paper describes the creation of a new CNN
and LSTM model architecture and classifying 8 annotation classes, along with using models
like ensemble algorithms, naive Bayes, support vector machines, and multi-layer perceptrons
to compare the proposed deep and recurrent neural networks. The issue of data imbalance is
countered by using data resampling techniques during data pre-processing.

\subsection{Machine Learning Interpretability}

The area that investigates the explainable properties of the ML models is called interpretability. It
determines if the model are black-boxes or if they have an explainable property. \cite{RN40}
defines interpretability as: ‘Interpretability is the degree to which a human can consistently predict the model’s result’. \cite{RN48} describes the need for interpretability of ML models in healthcare. The need
for explainable ’black-box’ models is important as it will allow the clinicians to make decisions
driven by data, which would in turn help patients receive personalized high-quality care. It will
improve efficiency in recognizing specific abnormalities help accelerate and optimize treatment.

\begin{figure}[h]
    \centering
    \includegraphics[width=0.5\textwidth]{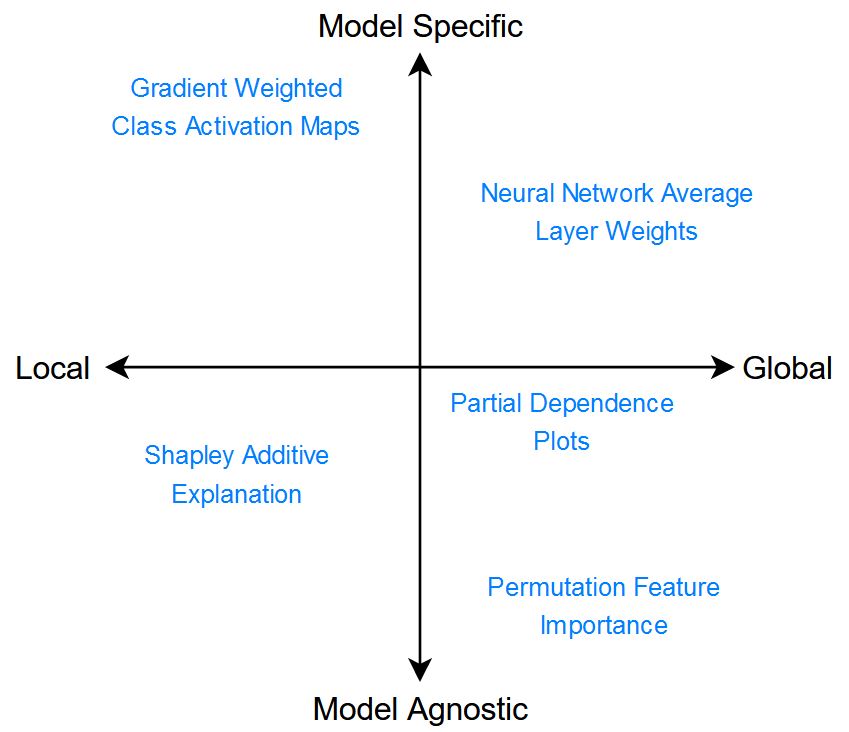}    \caption{A Guide to Interpretability}
    \label{fig:2}
\end{figure}

Interpretability can be divided into having four attributes namely: global and local interpretability, model-specific and model-agnostic interpretability. Figure \ref{fig:2} shows how different interpretability methods can have these attributes. As explained by \cite{RN40}, model-specific interpretability tools are confined and specific to one model and depend completely on the features and rules of that particular model. Model-agnostic tools, however, appear to be more applicable to post-hoc methods which can be used on any ML model by analysing input and
output feature weights. Global interpretability explains model decisions depending on conditional
interactions of the dependent and independent variables and features of the entire dataset, while
local interpretability seems to recognize prediction decisions for a single value in the data i.e., for the local region that the point is located at.

\cite{RN52} proposed a novel local interpretability technique of using Gradient-
weighted Class Activation Map (Grad-CAM) to visualize the saliency on both LSTM and
CNN models for ECG data. These saliency visualizations allowed the user to gain insight of
the predictions performed by the model. These interpretations, however, were performed on
individual time-series using beat-by-beat approach, which was unable to interpret and understand
the overall rules of the model’s decision on the entire dataset.

\cite{RN32} improved this ECG classification technique by applying saliency maps for
local interpretability on multiple ECGs of the same class to obtain an average explanation. This
allowed for a more general interpretation of the decisions made by the model, capturing all the
overarching rules. Their work involved segmentation of saliency maps into blocks for each beat
which allowed for quantitative comparison of decision-making process of the model. This was
implemented using the keras-vis library which has since been deprecated.

\section{Methodology}

\subsection{Data Pre-processing}

For reading the MIT-BIH dataset, the native python waveform-database (WFDB) package
was used, which is a library of tools for reading, writing, and processing WFDB signals and
annotations. This allowed us to obtain all the annotations for all normal and abnormal ECGs.

Referring back to table \ref{tab:1}, the annotation classes ’N’, ’L’, ’R’, ’V’, ’A’, ’F’, ’f’, ’/’ were used as most
of the ECG signals were assigned to these classes. There is an imbalance in the ECG dataset, where we see an abundance of ’N’ beats, while all the other beat classes do
not even pass the 10,000 threshold. This also only shows data from one of two channels of the
MIT-BIH database. In order to obtain all the beats, ECG signals from both the channels were
extracted and stacked. After extracting the 8 important classes that we will be working with and
removing all the other classes, we applied data clean-up processes. The first step was assigning
each of the 8 classes a number which would allow us to make the data purely numerical and easy
to work with. These assigned numerical values can be seen in table \ref{tab:1}.

Secondly, each patient record included a complete set of heartbeats, so each individual beat
was extracted from all the records by matching the R-peaks of the ECG with their respective
annotation class and appending the class’ numerical value at the end of the beat. To make every
beat contain equal amount of data points, the R-peaks were centered and equal amount of data
points were selected on both sides of the peak, this allowed us to maintain consistency for each
beat. To avoid irregular amplitudes between signals, all the beats were standardized. The patient record number was also appended at the end of the beat along with the annotation class number. All the cleaned-up data was stored into one .cvs file containing all beats of all records. The beat and patient hold-out methods were used for training and testing, split into 75-25\%, and 90-10\% respectively.

\begin{figure}[h]
    \centering
    \includegraphics[width=1\textwidth]{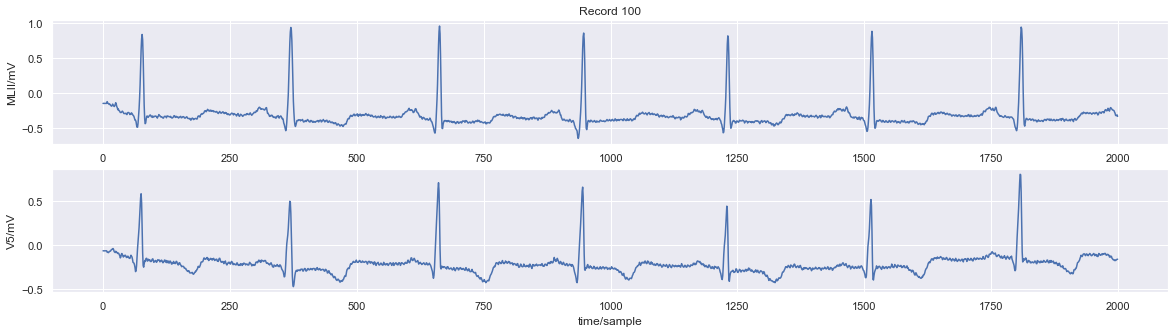}    \caption{Example of Continuous ECG Beats in a Record. (X-Axis: Timestamps, Y-Axis: Voltage)}
    \label{fig:3}
\end{figure}

\begin{figure}[h]
    \centering
    \includegraphics[width=0.3\textwidth]{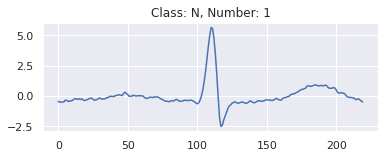}
    \includegraphics[width=0.3\textwidth]{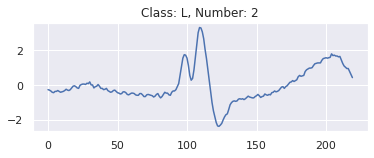} \includegraphics[width=0.3\textwidth]{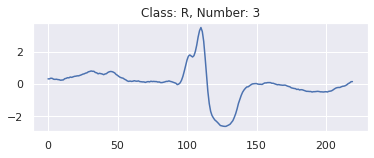} \includegraphics[width=0.3\textwidth]{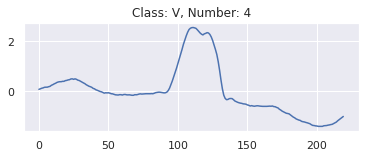} \includegraphics[width=0.3\textwidth]{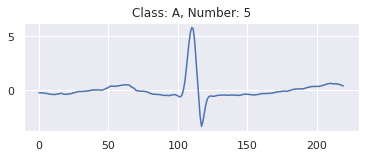} \includegraphics[width=0.3\textwidth]{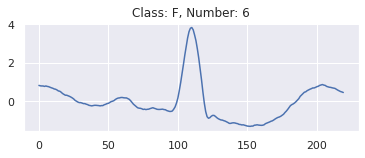} \includegraphics[width=0.3\textwidth]{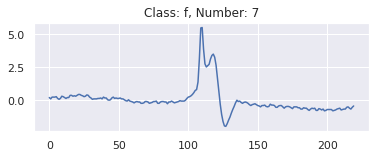}
    \includegraphics[width=0.3\textwidth]{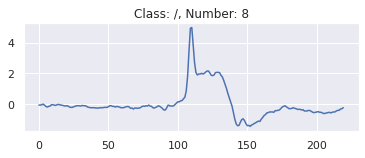} \caption{Examples of Single Beats of All Classes. (X-Axis: Timestamps, Y-Axis: Voltage)}
    \label{fig:4}
\end{figure}

Figure \ref{fig:3} shows an example of what the original data from MIT-BIH database looked like for
one patient in two channels, and as observed, the ECG signals are continuous and not standardized
between the two channels and are sampled at 360 Hz. However, figure \ref{fig:4} shows examples of extracted ECG signals with the annotation classes after data pre-processing, allowing us to see
the standardization, and r-peak centering implemented on the data points.

\subsection{Data Resampling and Visualization}

\begin{figure}[h]
    \centering
    \includegraphics[width=0.49\textwidth]{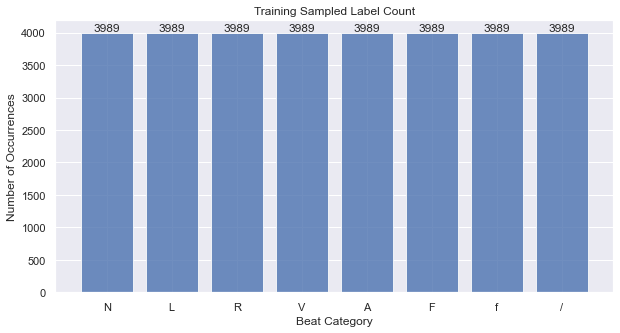}
    \includegraphics[width=0.49\textwidth]{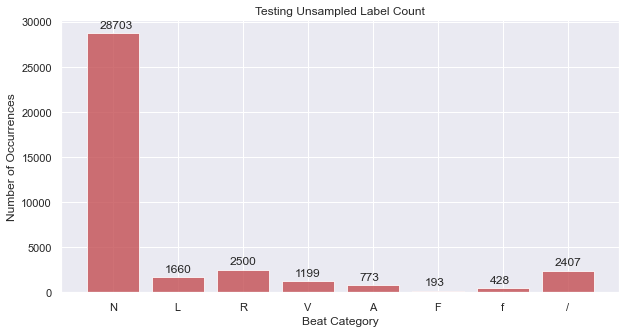}
    \caption{Sampled Train and Test Dataset for Beat Holdout 75/25}
    \label{fig:5}
\end{figure}

\begin{figure}[h]
    \centering
    \includegraphics[width=0.49\textwidth]{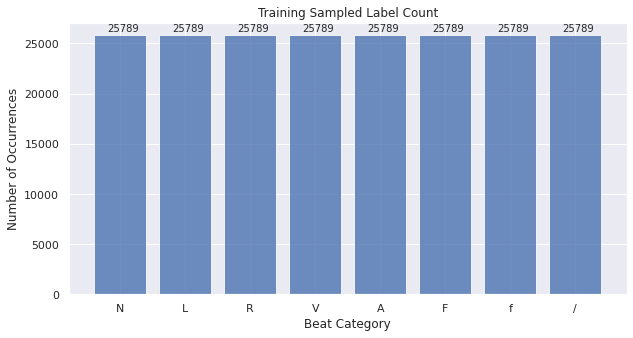}
    \includegraphics[width=0.49\textwidth]{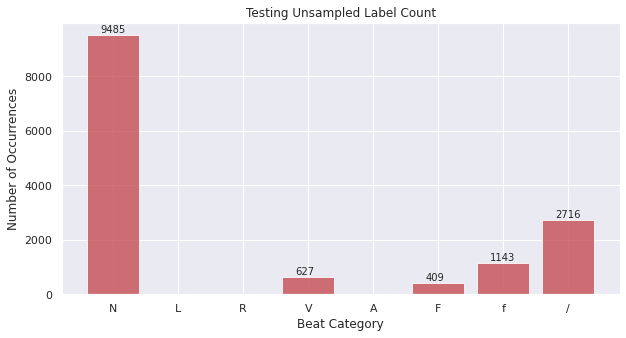}
    \caption{Sampled Train and Test Dataset for Patient Holdout 75/25}
    \label{fig:6}
\end{figure}

To address the imbalance between the classes in the MIT-BIH dataset, we used the resample technique by Sci-kit Learn \cite{RN44}. This resampling method uses bootstrap method which estimates statistics on a data population by sampling a dataset with replacement through iteration, using a sample size and the number of repeats. For up-sampling and down-sampling, the $n_{-}$samples value was calculated by taking the mean values of the total number of beats of the abnormal classes. Figures \ref{fig:5} and \ref{fig:6} show beats with their respective annotation classes from both ECG channels for beat and patient holdout splits. We observe that after resampling, all the 8 classes in the train dataset have 3989 samples for the beat holdout method and 25789 samples for the patient holdout method.

\begin{figure}[h]
    \centering
    \includegraphics[width=0.29\textwidth]{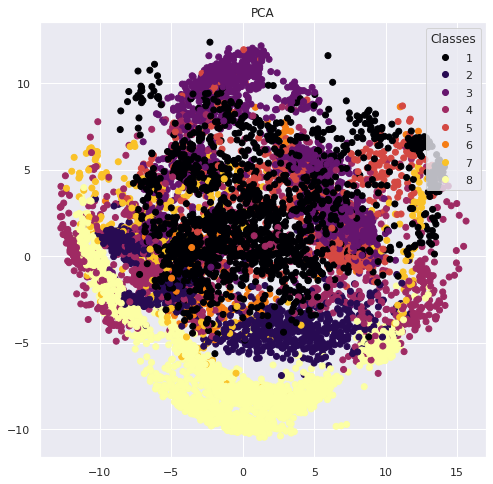}
    \includegraphics[width=0.29\textwidth]{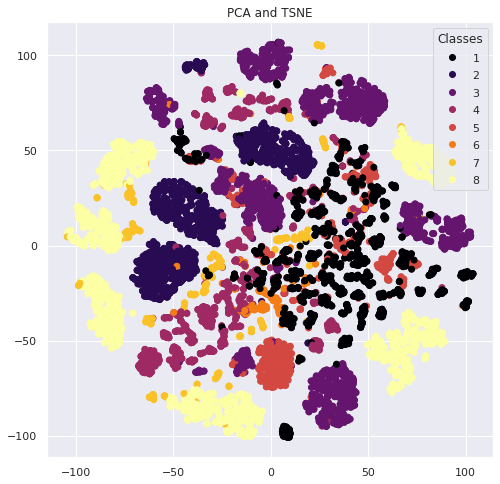}
    \includegraphics[width=0.29\textwidth]{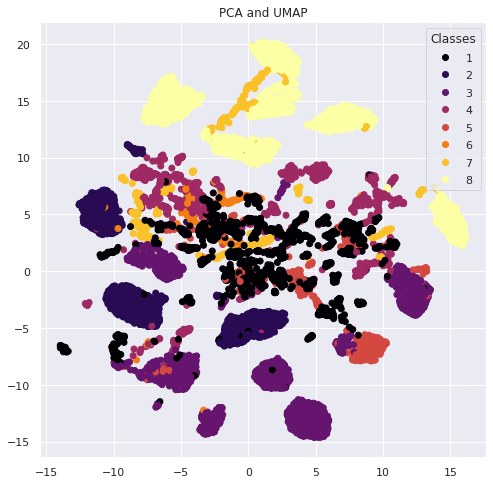}
    \caption{Clustering Unsampled Data per Class}
    \label{fig:7}
\end{figure}

\begin{figure}[h]
    \centering
    \includegraphics[width=0.29\textwidth]{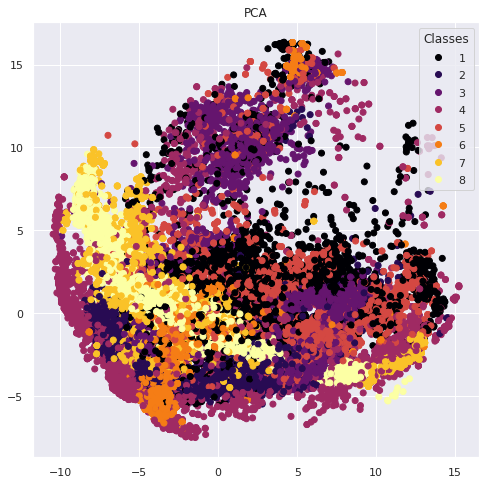}
    \includegraphics[width=0.29\textwidth]{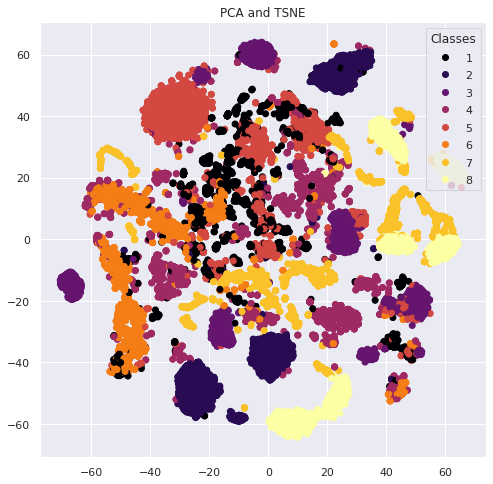}
    \includegraphics[width=0.29\textwidth]{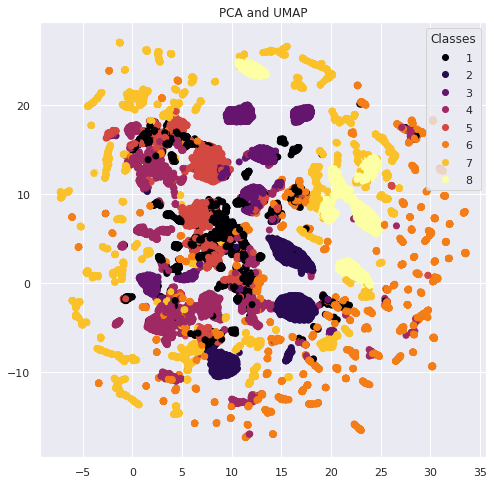}
    \caption{Clustering Sampled Data per Class}
    \label{fig:8}
\end{figure}

Dimensionality reduction and data visualization techniques were implemented to visualize the ECG classes, since clustering using unsupervised learning is helpful at visualizing different classes and shows us how similar or different the classes behave for unsampled and sampled datasets. These visualization techniques were performed on both unsampled \ref{fig:7} and sampled \ref{fig:8} datasets to understand how close the bootstrap resampling data is clustered to the unsampled data clusters. 

Principal Component Analysis (PCA) \cite{RN31} was performed on both datasets for dimensionality reduction with $n\_components$ 50 . We see in figures, a cloud of data points of different classes but clear clusters were not visible. To improve upon PCA, t-distributed Stochastic Neighbor Embedding (t-SNE) \cite{RN50} with $n\_components$ set to 2 was implemented on PCA. As observed in the sampled figure, it shows us big clusters of different classes, but the data was still fairly spread out. However, for TSNE good clustering was noticed for classes L, R, A and /. To try to further improve this, Uniform Manifold Approximation and Projection (UMAP) \cite{RN39} was implemented on PCA, with $n\_components$ 2 and $n\_neighbors$ 80. As seen in the UMAP figure, the bigger clusters disappear and the data lumps into a balls of smaller clusters which is not very useful in visualizing different classes, but for the unsampled figure, good clustering was noticed for classes $\mathrm{N}, \mathrm{F}, \mathrm{f}$. These small data lumps can be due to the upsampled beats created by bootstrap resampling techniques which have similar time-series data points.

\subsection{Machine Learning Models}

This paper uses various ML models, specifically Gradient Boosting (GBC), ADA Boosting (ADA),
Random Forest (RFC), Gaussian Naive Bayes (NB), Neural Network Multi-Layer Perceptron
(NNMLP), C-Support Vector (SVC), Convolutional Neural Network (CNN), and Long Short-
Term Memory Network (LSTM). These eight ML classifier models range from ensemble-based
algorithms, Bayes algorithm, support vector machine, to deep and recurrent neural networks.
This diversified selection of models allows us to investigate and evaluate their performance on
multi-class classification of uni-variate time-series ECG data and interpret the best performing
models.

\subsubsection{Model Architectures and Hyper-parameters}

\begin{table}[!htb]
    \begin{minipage}{.5\linewidth}
      \caption{Layers of CNN Model}
      \centering
        \begin{tabular}{lccc}
        \hline \textbf{Layer} & \textbf{Filters} & \textbf{Size} & \textbf{Activation} \\
        \hline 1D-Conv & 128 & 16 & ReLU \\
        BatchNorm & $-$ & $-$ & $-$ \\
        1D-Conv & 32 & 16 & ReLU \\
        BatchNorm & $-$ & $-$ & $-$ \\
        1D-Conv & 9 & 16 & ReLU \\
        1D-MaxPoot & $-$ & 2 & $-$ \\
        Flatten & $-$ & $-$ & $-$ \\
        Dense & 512 & $-$ & ReLU \\
        Dense & 128 & $-$ & ReLU \\
        Dense & 32 & $-$ & ReLU \\
        Dense & 9 & $-$ & Softmax \\
        \hline
        \label{tab:2}
        \end{tabular}
    \end{minipage}%
    \begin{minipage}{.5\linewidth}
        \caption{Layers of LSTM Model}
      \centering
        \begin{tabular}{lccc}
        \hline \textbf{Layer} & \textbf{Filters} & \textbf{Size} & \textbf{Activation} \\
        \hline LSTM & 128 & 1 & $-$ \\
        LSTM & 9 & $-$ & $-$ \\
        1D-MaxPool & $-$ & 2 & $-$ \\
        Flatten & $-$ & $-$ & $-$ \\
        Dense & 512 & $-$ & ReLU \\
        Dense & 128 & $-$ & ReLU \\
        Dense & 32 & $-$ & ReLU \\
        Dense & 9 & $-$ & Softmax \\
        \hline
        \label{tab:3}
        \end{tabular}
    \end{minipage} 
\end{table}

The CNN model architecture proposed and investigated in this paper is a 11 weighted layer classifier. As seen in table \ref{tab:2}, the first four layers of my sequential model are a combination of two 1D-convolutional layers and two batch normalization layer pairs, with kernel size 16 , filters 128,32 , and activation ReLU for the convolutional layer. The next two layers are the final 1D-convolutional layer with filters 9 and 1D-max pooling layer with pool size 2 . The output of the max pooling layer is flattened by a flatten layer and the data is then used as input for the final four fully connected dense layers, where the first three are ReLU activated, and the final layer uses softmax activation for the final output. The model is compiled using the adaptive moment estimation (Adam) optimizer, with the default learning rate of $0.001$, and categorical cross-entropy loss function. The model was fit on the training dataset for 10 epochs and 64 batch size.

The LSTM classifier model proposed and investigated in this paper consisted of 8 weighted layers. As seen in table \ref{tab:3}, the first two layers were LSTM layers with filter 128 and 9, followed by 1D-max pooling layer with pool size 2 . Like the CNN, the max pooling layer's output is flattened by a flatten layer and then is inputted into four fully connected dense layers with activation ReLU for the first three and softmax for the final layer. The model is compiled using the Adam optimizer, with the default learning rate of $0.001$, and categorical cross-entropy loss function. The model was fit on the training dataset for 10 epochs and 256 batch size.

The hyper-parameter for the rest of the models were as follows: both GBC and ADA used $n\_estimators$ $=100$, RFC had $n\_estimator$ $=10$ with $max\_depth$ $=10$, NNMLP had $max\_iter$ $=100$ , and NB and SVC used their default parameters.

\subsection{Interpretability Techniques}

\begin{figure}[h]
    \centering
    \includegraphics[width=0.7\textwidth]{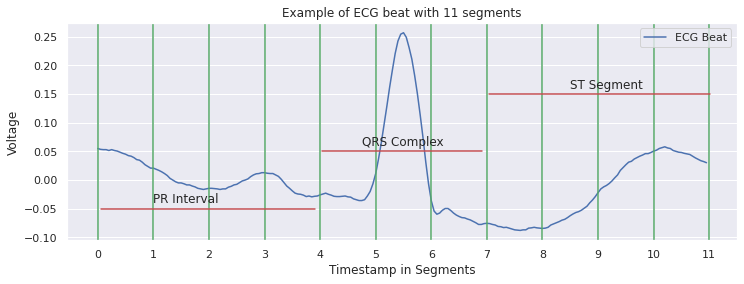}
    \caption{ECG Beat Sliced into 11 Segments. (X-Axis: Timestamps, Y-Axis: Voltage)}
    \label{fig:9}
\end{figure}

Once we have the models, we need a way to interpret them, so we can compare the behaviour
of each. This section discusses how we do this. The ECG beats were divided into slices of 11
segments which would allow us to understand and interpret which segment is being given more
importance by the model during classification. The slices were made by replacing the data points
with the average point for each slice. Figure \ref{fig:9} shows a sample ECG beat sliced into 11 segments.
Here the segments 1-4 cover the PR interval, the next three segments 5-7 cover the QRS complex
and the final four segments 8-11 cover the ST segment. This plot is extremely important as the
segment numbers and the names of these group of segments will be frequently mentioned in
this entire section. We expected to see the model focusing on important morphological features
of the ECG beat, such as the PR interval, the QRS complex, and the ST segment. It could also
focus on areas we do not expect due to the capacity of these methods to pick up key features and
interactions between pixels in an input image.

\subsubsection{Partial Dependency Plots (PDP)}

In PDP we expected to see the dependence between the target response and a set of input features
of interest, marginalizing over the values of other complementary features.

Two-way PDP were plotted per class and one-way PDP per ECG segment for the NNMLP
model, where the normal class was compared against different abnormal classes. For global
interpretability, the features are considered to be classes and slices of ECG beats for the complete
dataset. The two-way class plots showed that N class as no marginal effect on itself when NNMLP
makes its decisions, but every other abnormal class seems to be affected differently by the N
class. The per class PDP can be inaccurate as, the practical highest number of features in a partial
dependence function is two which is not the fault of PDP, but of the 2-dimensional representation
and the inability to comprehend more than 3 dimensions. Since classes are not really features, we
move on to one-way PDP to explain the NNMLP model per segment.

The one-way plot per segment showed that segments 1 and 5 have a similar dependence curve.
This tells us that this pair of 1 and 5 segments have the same marginal effect on NNMLP’s decision-
making process. As for other segments, each of them shows a different partial dependence curve
which means that their marginal effect on the predicted outcome of NNMLP was unique, but
this did not allow us to interpret the model’s predictions. The relevant graphs can be found in
the github repository.

This interpretability technique did not allow us to clearly understand the dependence between
features for most segments and had no scope to be quantitatively analysed. Since PDP provide
an extremely global interpretation, we move on to techniques which would allow us to further
investigate interpretability in a deeper aspect.

\subsubsection{Shapley Additive Explanations (SHAP)}

With SHAP we expected a visual explanation of the model by using shapley values to calculate
the importance of a feature by comparing what a model predicts with and without the feature.
The desired result would be to obtain ECG segments considered important by the SVC model
in each class.

KernelSHAP technique was implemented to calculate feature importance for the SVC model
as it does not support TensorFlow models on uni-variate time-series data yet. SHAP summary
plot was calculated for ECG segments which allowed us to see the behaviour of each segment for
every class. SHAP treats the segments of the ECG as features and explains the model’s decision by
giving it a SHAP value, so the 11 segments of the ECG beats are shown as features 0 through 10.

\begin{figure}[h]
\centering
\begin{minipage}{.5\textwidth}
  \centering
  \includegraphics[width=1\textwidth]{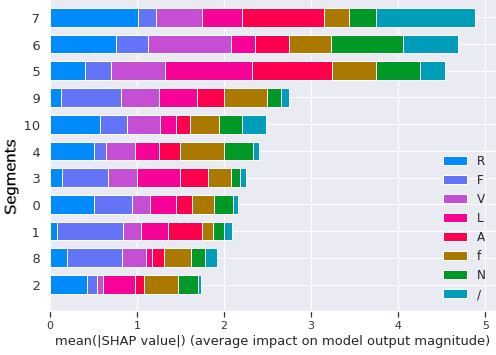}
  \captionof{figure}{SHAP Summary Plot}
  \label{fig:10}
\end{minipage}%
\hfill
\begin{minipage}{.5\textwidth}
  \centering
  \includegraphics[width=1\textwidth]{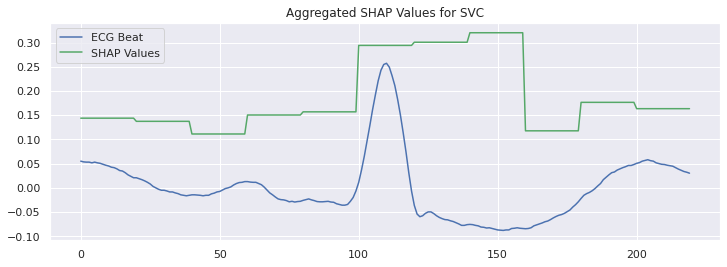}
  \captionof{figure}{Global Aggregate of SHAP values. (X-Axis:
Timestamps, Y-Axis: Scaled Values)}
  \label{fig:11}
\end{minipage}
\end{figure}

In figure \ref{fig:10}, the summary plots show the mean of the SHAP values i.e the average impact the
features and classes had on the magnitude of the model’s outcome. Here we see that features
5, 6, 7 were the most important segments overall but the effect of each individual class can be
noticed to be different. The figure \ref{fig:11} takes the global aggregate of the SHAP values and displays
them on top of a sample ECG beat. We observe that segments 5, 6, 7 are clearly considered more
important than the other segments in a global visualization, which represents the QRS complex.
Local visualizations were not possible as SHAP explained only a fraction of instances of the test
data, which is not enough to give reliable results.

These interpretability techniques explain a lot about the model’s decision-making process and
gives accurate explanations, but KernelSHAP is extremely time consuming and interprets only
the first 50 instances of the data, making it not fit for explaining deep and recurrent neural
networks for time-series data. An alternative for the feature importance method used in SHAP is
permutation feature importance, which is explored in the next sub-section.

\subsubsection{Permutation Feature Importance (PFI)}

In PFI we expected to see importance weights assigned to different ECG segments and which
segment each model considers the most important. This technique works for SK-Learn and TensorFlow (using a surrogate model) models which allows us to compare these feature importance
scores between each of the classifiers. These values can be quantitatively analysed making PFI
easier to evaluate.

The ELI5 library, which allows to explain weights and predictions of a classifier,
was used for the computation of PFI. This library provides a way to compute feature importance
for any black-box estimator by measuring how score decreases when a feature is unavailable. More figures can be found on github, where the top three segments (features) in most classifiers are 5, 6, 7. which represent the QRS complex in the ECG beats, shown in figure \ref{fig:9}. The interpretation is global since we
are looking at the complete data as opposed to a since class or a single beat. This interpretation,
however global, allows us to consistently explain the predictions of all the models. This method
proves to be more helpful as unlike SHAP and PDP methods, it works on all classifiers.

\begin{figure}[h]
    \centering
    \includegraphics[width=0.49\textwidth]{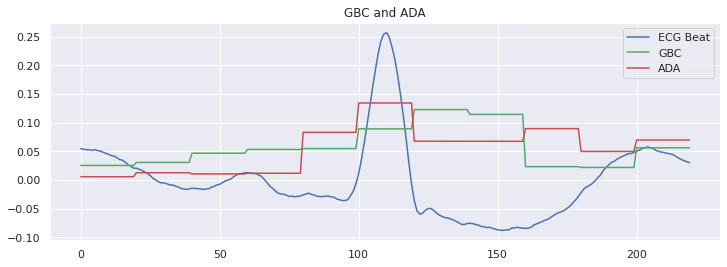}
    \includegraphics[width=0.49\textwidth]{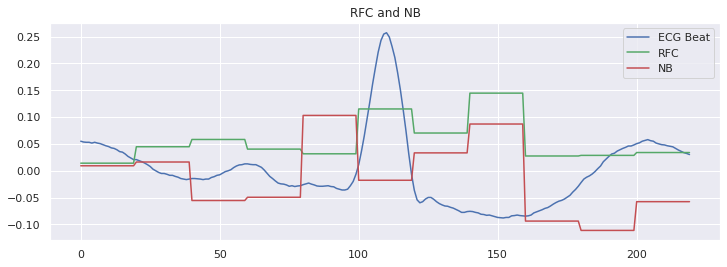}
    \includegraphics[width=0.49\textwidth]{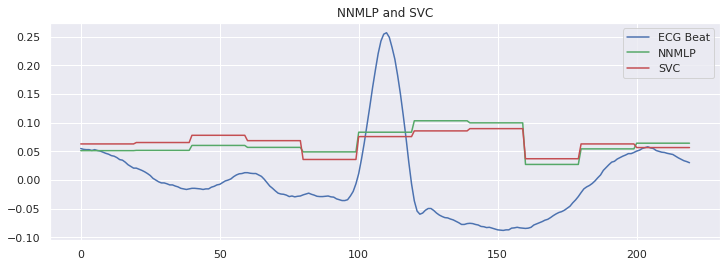}
    \includegraphics[width=0.49\textwidth]{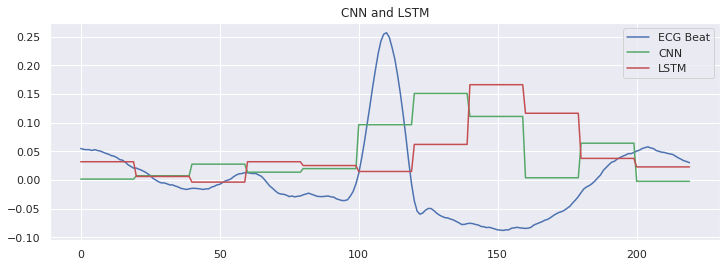}
    \caption{Feature Importance of all Models on Sample ECG Beat. (X-Axis: Timestamps, Y-Axis: Scaled Values)}
    \label{fig:12}
\end{figure}

In figure \ref{fig:12}, the feature importance scores are represented on a sample ECG beat to better
visualize the explanation of this method for each classifier. It can be observed that GBC, RFC,
NNMLP, SVC, and CNN give the QRS complex of the ECG beat the highest feature importance.
For ADA and NB, these classifiers give lesser importance to the QRS complex but still higher
than other segments. The LSTM classifier gives the ST segment higher importance than the
QRS complex. However, these results include all the classes of the classified beats which is a very
global interpretation of the explanation of the model’s decision. Since we cannot obtain the PFI
per class, the classified beats were separated into correctly and the incorrectly classified beats and implement feature importance on them.

\begin{figure}[h]
    \centering
    \includegraphics[width=0.49\textwidth]{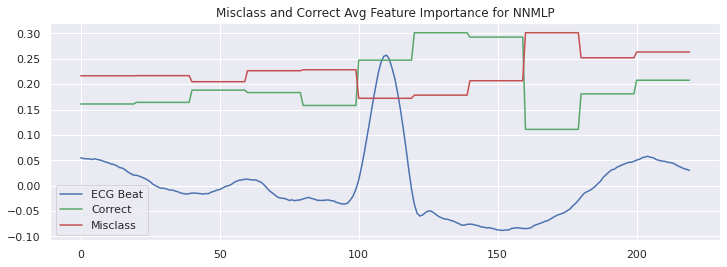}
    \includegraphics[width=0.49\textwidth]{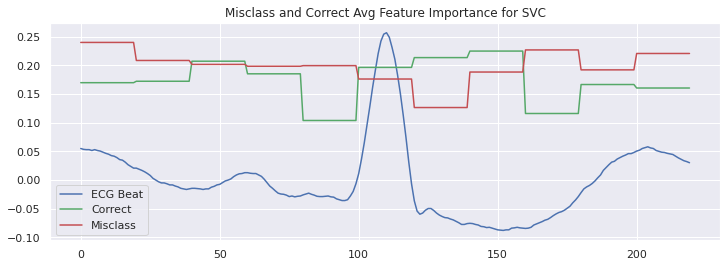}
    \caption{Feature Importance of Correctly and Incorrectly Classified Beats on Sample ECG Beat. (X-Axis: Timestamps, Y-Axis: Scaled Values)}
    \label{fig:13}
\end{figure}

Figure \ref{fig:13} shows the representation of PFI weights per segment for the correctly classified
and mis-classified beats for NNMLP and SVC models on a sample ECG beat. We notice that
the feature importance representation of the mis-classified beats (in red) does not clearly show
important ECG segments. The correct segments (in green) however, show that the QRS complex
is given higher importance weights compared to the other segments. We would expect the
model to look at the QRS complex, and so the correctly classified beat plot makes sense. The fact
that with mis-classified beats the model is looking everywhere, suggests that the model does not
know where to look.

PFI gave us quantitative values of the prediction of the models per segment, this can be considered
a global explanation of the model as it focuses on the entire dataset and the drawback of global interpretation techniques is that we cannot look at explanation of specific beats and classes. Grad-CAM will explore a more local interpretation of each beat along with the correct and incorrectly classified beats for the TensorFlow CNN and LSTM models in the next section.

\subsubsection{Gradient-Weighted Class Activation Maps (Grad-CAM)}

In the Grad-CAM implementation we expected to see a class-specific saliency heat-map of the
ECG beats. The higher Grad-CAM values will indicate the data points and segments that the
layers of the models give higher gradient weights to, which means that they are more important.
The darker and lighter segments of the heat-map will indicate higher and lower Grad-CAM
values, respectively.

\begin{figure}[h]
    \centering
    \includegraphics[width=0.59\textwidth, trim=0 0 0 0, clip]{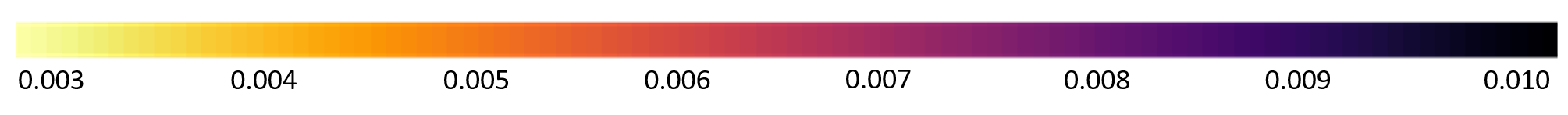}
    
    \begin{subfigure}{.49\linewidth}
        \includegraphics[width=0.99\linewidth, trim=0 0 100 0, clip]{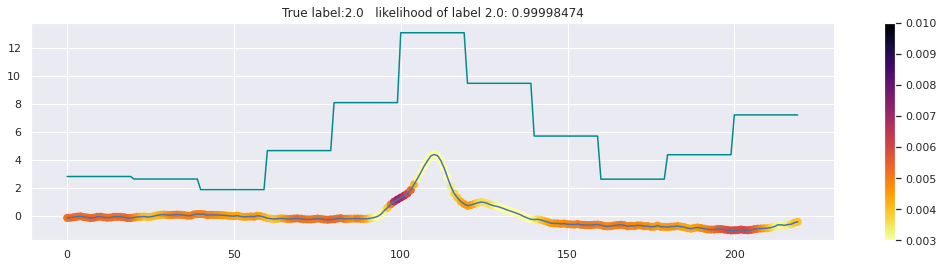}
        \includegraphics[width=0.99\linewidth, trim=0 0 100 0, clip]{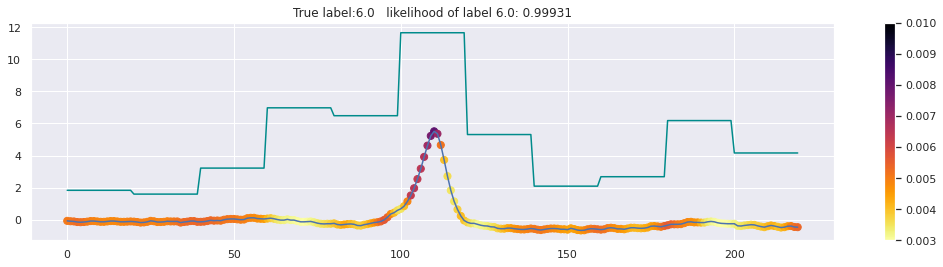}
        \caption{Correctly Classified Beats}
    \end{subfigure}
    \hfill
    \begin{subfigure}{.49\linewidth}
        \includegraphics[width=0.99\linewidth, trim=0 0 100 0, clip]{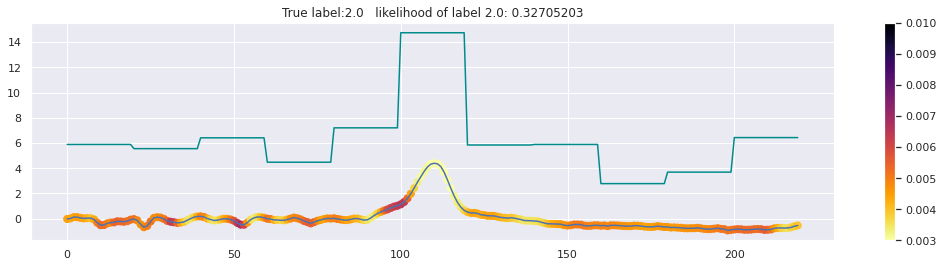}
        \includegraphics[width=0.99\linewidth, trim=0 0 100 0, clip]{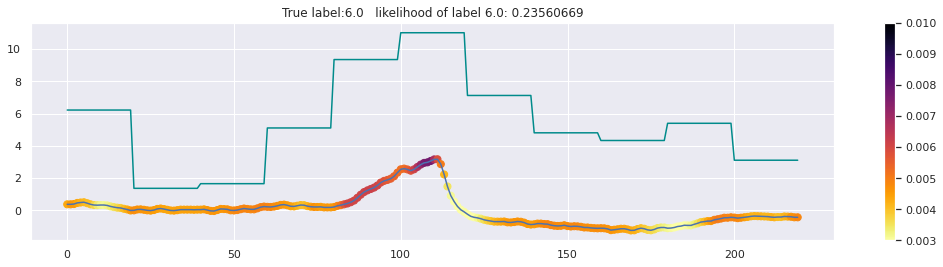}
        \caption{Incorrectly Classified Beats}
    \end{subfigure}

    \caption{Grad-CAM (heat-map) and Layer Weights (in green) per L (top) and F (bottom) Class for CNN. (X-Axis: Timestamps, Y-Axis: Scaled Values)}
    \label{fig:14}
\end{figure}

\begin{figure}[h]
    \centering
    \includegraphics[width=0.59\textwidth, trim=0 0 0 0, clip]{images/metrics/colorbar_gradcam.png}
    
    \begin{subfigure}{.49\linewidth}
    \includegraphics[width=0.99\textwidth, trim=0 0 100 0, clip]{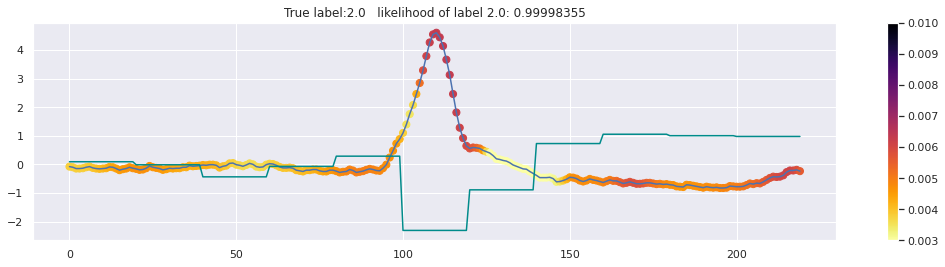}
    \includegraphics[width=0.99\textwidth, trim=0 0 100 0, clip]{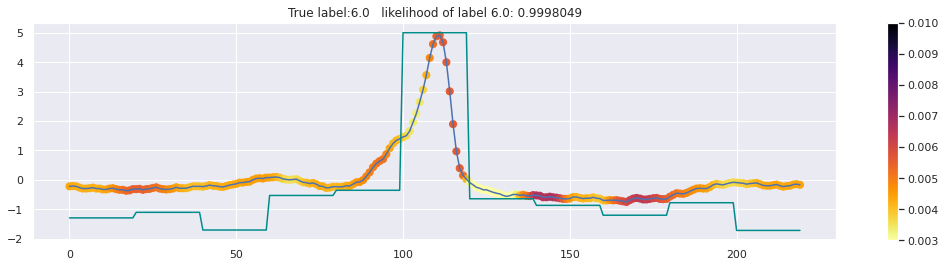}
    \caption{Correctly Classified Beats}
    \end{subfigure}
    \hfill
    \begin{subfigure}{.49\linewidth}
    \includegraphics[width=0.99\textwidth, trim=0 0 100 0, clip]{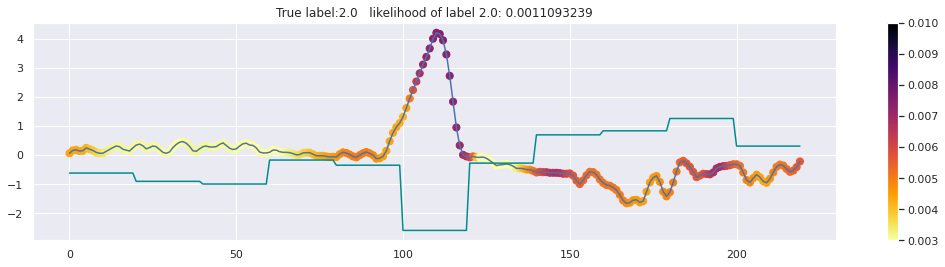}
    \includegraphics[width=0.99\textwidth, trim=0 0 100 0, clip]{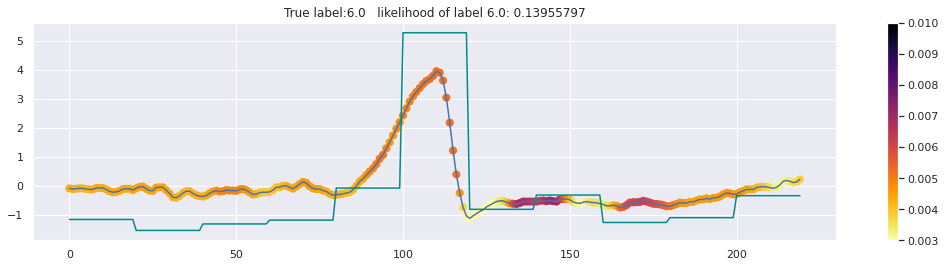}
    \caption{Incorrectly Classified Beats}
    \end{subfigure}
    
    \caption{Grad-CAM (heat-map) and Layer Weights (in green) per L (top) and F (bottom) Class for LSTM. (X-Axis: Timestamps, Y-Axis: Scaled Values)}
    \label{fig:15}
\end{figure}

Grad-CAM for time-series saliency was implemented on the final convolutional layer for the
CNN model and the final LSTM layer for the LSTM model. Using the keras back-end function, the weights for these finals layers as well the softmax layer were obtained and the CAM formula
2.4 was implemented on the values. This gave us the activation for the final layers and the
obtained values were plotted using a scatter plot on the ECG beats. We visualize in figures \ref{fig:14}
and \ref{fig:15} that the discriminating regions of the time-series for the correct and incorrect classes
are highlighted. The title indicates the arrhythmia beat class label and the probability values of
predicted class. To allow us to visualize which ECG segment is being deemed the most important
by the classifiers, the layer weight values were averaged for each segment and plotted in green.
The scale provided on top of the plots indicate the range Grad-CAM values of the final layers.

Figure \ref{fig:14} and \ref{fig:15} shows the saliency maps for an individual beat of classes L and F for CNN
and LSTM models, respectively. We see that in figure \ref{fig:14} the QRS complex is being considered the most important by the CNN model for the classification
process. Although in figure \ref{fig:15}, it is not clear which segment is deemed as more important by
the LSTM model, the Grad-CAM values are discriminating the QRS complex way more than
the CNN model, as the R-peaks can be seen having darker scatter points than the PR interval.

\begin{figure}[h]
    \centering
    \begin{subfigure}{.49\linewidth}
    \includegraphics[width=0.99\textwidth, trim=0 0 0 0, clip]{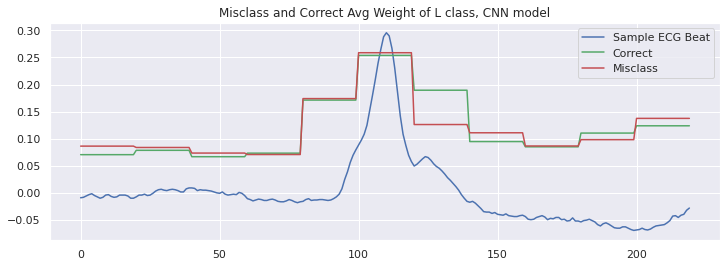}
    \includegraphics[width=0.99\textwidth, trim=0 0 0 0, clip]{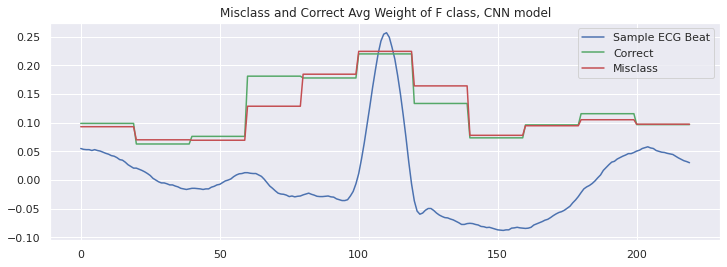}
    \caption{CNN Model}
    \end{subfigure}
    \hfill
    \begin{subfigure}{.49\linewidth}
    \includegraphics[width=0.99\textwidth, trim=0 0 0 0, clip]{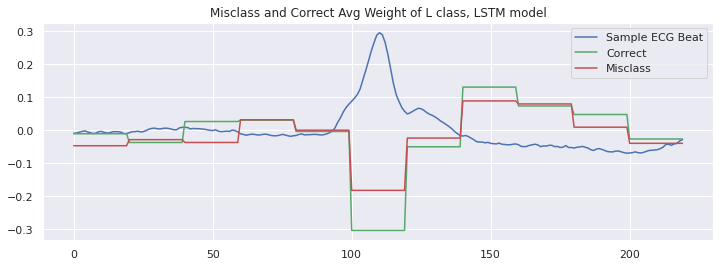}
    \includegraphics[width=0.99\textwidth, trim=0 0 0 0, clip]{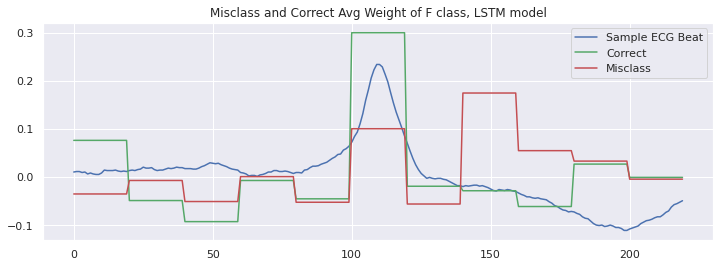}
    \caption{LSTM Model}
    \end{subfigure}
    \caption{Average Layer Weights per L (top) and F (bottom) Class for CNN and LSTM. (X-Axis: Timestamps, Y-Axis: Scaled Values)}
    \label{fig:16}
\end{figure}

More figures on github show the average layer weights of all correctly and incorrectly classified beats for
CNN and LSTM model, respectively. The models, on average, are looking at the segments of
both correct and incorrectly classified beats equally. This however is a very global interpretation
and does not actually tell us about the model’s behaviour per class. To visualize a more local
interpretation of the models, the average weights for L and F classes are seen in figure \ref{fig:16}. Here
it is observed that the CNN model is giving higher importance to the QRS complex for both
L and F classes, while the LSTM model focuses more on the ST segments while making the
classification decisions. This observation is supported by the PFI weights for CNN and LSTM
from figure \ref{fig:12}. Plots for rest of the classes can be found on github, where we
can locally interpret the explanation of both the models’ and compare these explanations between
all eight classes.

\begin{figure}[h]
    \centering
    \begin{subfigure}{.49\linewidth}
    \includegraphics[width=0.99\textwidth, trim=0 0 0 0, clip]{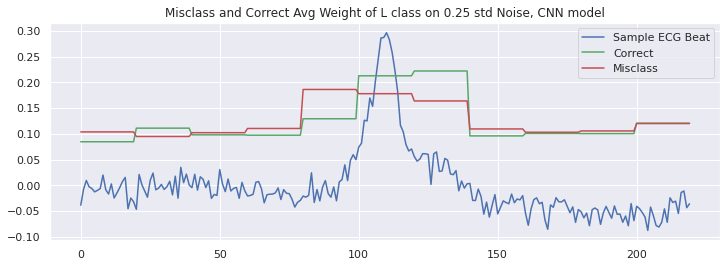}
    \includegraphics[width=0.99\textwidth, trim=0 0 0 0, clip]{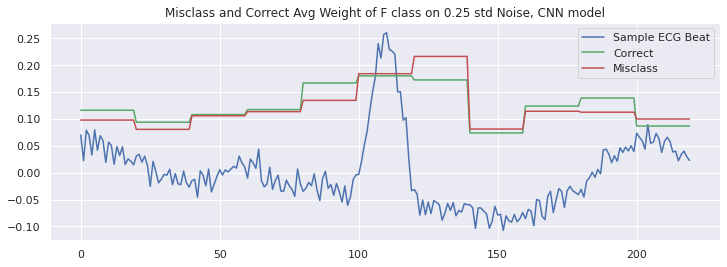}
    \caption{CNN Model}
    \end{subfigure}
    \hfill
    \begin{subfigure}{.49\linewidth}
    \includegraphics[width=0.99\textwidth, trim=0 0 0 0, clip]{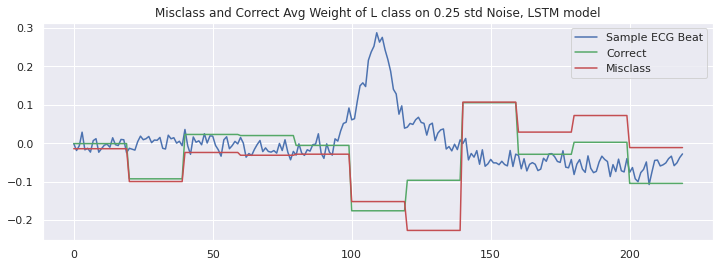}
    \includegraphics[width=0.99\textwidth, trim=0 0 0 0, clip]{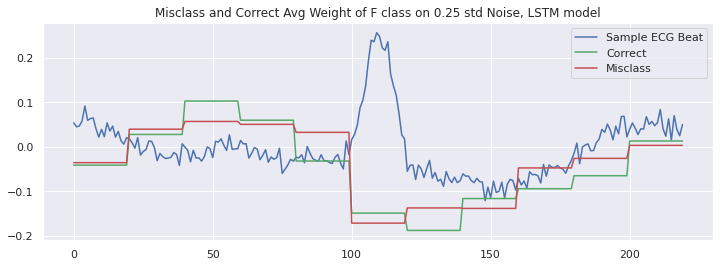}
    \caption{LSTM Model}
    \end{subfigure}
    \caption{Average Layer Weights per L (top) and F (bottom) Class for std 0.25 Noise on Test Data for CNN and LSTM. (X-Axis: Timestamps, Y-Axis: Scaled Values)}
    \label{fig:17}
\end{figure}

To examine the robustness of Grad-CAM and layer weight interpretability maps, slightly different
test data was used by introducing Gaussian noise and examining the correctly and incorrectly
classified samples. The distributions of the noise were created by taking standard deviation of the
test data. Low frequency noise, anything less than about 20Hz can be caused by a patient moving
when the ECG is being taken and high frequency noise can be caused by a faulty electrode or
general electrical noise \cite{RN35}. To mimic these scenarios, Gaussian noise of std 0.25 was
introduced in the test data. In figure \ref{fig:17} we see that the CNN model is still focusing on the QRS complex in both L and F classes for classification, while the LSTM model seems to be focusing
on the ST segment in L class and the PR interval in the F class.

This allows us to conclude that introducing noise in the dataset does not completely weaken the
ability of the classifiers in making predictions. Although the important segments are not clearly
being distinguished for the LSTM model, the CNN model still focuses on the same segments it
was looking at in the data without noise. The robustness of the interpretability is also verified, as
the effective segments of the layer weights in figure \ref{fig:17} mostly agree with figure \ref{fig:16}, which
means that the models dealt with noisy data pretty well. Although their performance may vary
on completely new datasets with ECG beats obtained from different patients.

To conclude this subsection, we observed that Grad-CAM was able to interpret the ECG signals
locally and globally. This method explained the predictions of both convolutional and recurrent
neural networks at each time-series data point per beat and also allowed us to understand the
overall behaviour of the model. The models mostly gravitated towards the QRS complex being
the most important segment while making predictions, although it did differ slightly depending
on the type of class. Classes L and F were used for Grad-CAM explanation in the main text, and
explanation of the rest of the classes can be found on github.

\subsubsection{Summary}

The methodology chapter explained the data pre-processing and visualization techniques, along
with mentioning the architectures and hyper-parameters of different models used in this ex-
periment. Since most interpretability libraries do not support time-series data, the ECG beats
were sliced into 11 segments where different group of segments represented the morphology
of the ECG beat i.e., the PR interval, QRS complex, and the ST segment. The implementation
process of the four interpretability techniques used in this investigation was described. These
interpretability methods were performed on the first fold of the K-fold cross validation evaluation
of the model. PDP turned out to be extremely uninformative at explaining the NNMLP model,
while SHAP on SVC was extremely time consuming and explained a fraction of instances of
the dataset. PFI being the only model agnostic technique, gave global explanations of all models.
Grad-CAM gave both local and global explanations of deep learning CNN and LSTM models.

\section{Results and Evaluation}

The models were running on Google Co-laboratory notebooks utilizing the GPU. After training and fitting the models on the dataset, performance metrics were calculated which included precision-recall, F1-scores, and accuracy for each class and for the complete dataset as well. The documentation \cite{RN44} describes precision-recall as: 'In information retrieval, precision is a measure of result relevancy, while recall is a measure of how many truly relevant results are returned.', F1 scores as: 'The F1 score can be interpreted as a weighted average of the precision and recall, where an $F 1$ score reaches its best value at 1 and worst score at 0 .' and accuracy as: 'The set of labels predicted for a sample must exactly match the corresponding set of labels in target.'

\subsection{K-Fold Cross-Validation (Holdout Beats)}

\begin{figure}[h]
    \centering
    \includegraphics[width=0.8\textwidth]{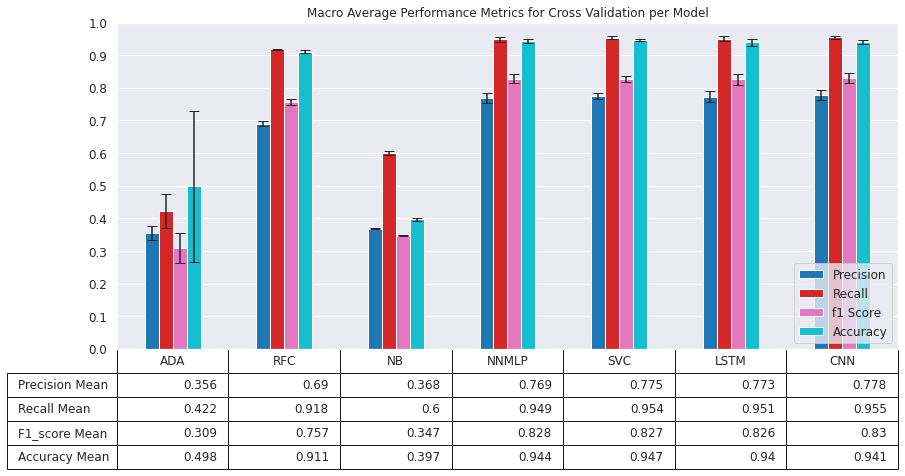}
    \caption{Macro Average Performance Metrics per Model on K-fold Cross-Validation. (X-Axis: Models, Y-Axis: Scores)}
    \label{fig:18}
\end{figure}

Cross-validation is used to quantitatively estimate and compare the performance of an ML model
on unseen data to estimate the model’s general performance. Stratified K-Fold cross-validation
was performed on the complete unsampled dataset for all the classifier models with 6 folds, which
allowed the given data to be split into 6 different groups for testing and training purposes where
the distribution of classes is maintained in the test and train data. The K-Fold cross-validation
shuffles the dataset randomly and splits it into k groups, and from each group it obtains a group
as hold out while using the rest of it for training.

\subsubsection{Classification Performance Metrics}

The model is evaluated, and the evaluation score is retained, and the model is reset for the next
fold. Each observation in the data is assigned to a specific group and remains in that group for the
duration of the procedure, which allows each sample to be used as hold out once and to train k-1
times. The results of the K-Fold cross-validation are summarized with the mean metric score
with the standard deviation between the 6 folds, the metrics are mentioned in figure \ref{fig:18}. Since
gradient boosting classifier in SK-Learn does not support multi-threading, 6 fold cross-validation
was taking more than 4 hours to compute, so it was discarded.
The macro average score does not consider the proportions of each label in the dataset, while
the weighted average performance metrics considers the number of instances of each class while
calculating the results, this allows for the metrics to account for imbalance in classes, and results for these can be found on github. Note that interpretability techniques were performed on the models’ first cross-validation fold.

\begin{figure}[h]
    \centering
    \includegraphics[width=0.3\textwidth, trim=0 0 60 0, clip]{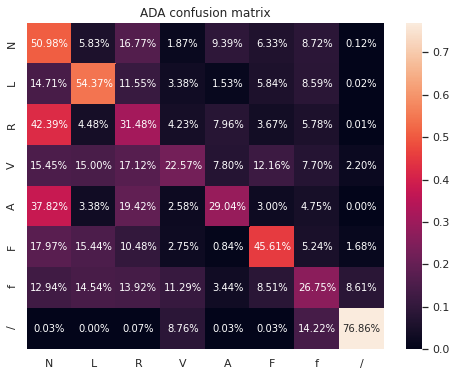}
    \includegraphics[width=0.3\textwidth, trim=0 0 60 0, clip]{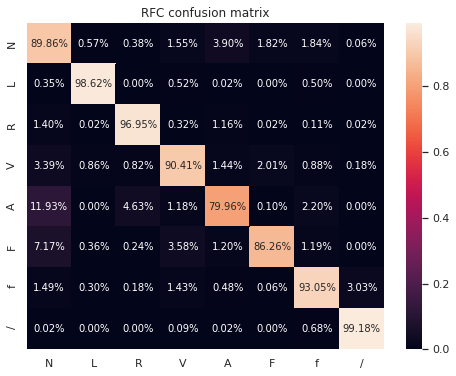}
    \includegraphics[width=0.3\textwidth, trim=0 0 60 0, clip]{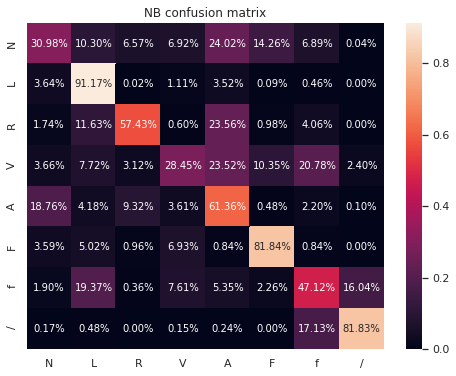}
    \includegraphics[width=0.3\textwidth, trim=0 0 60 0, clip]{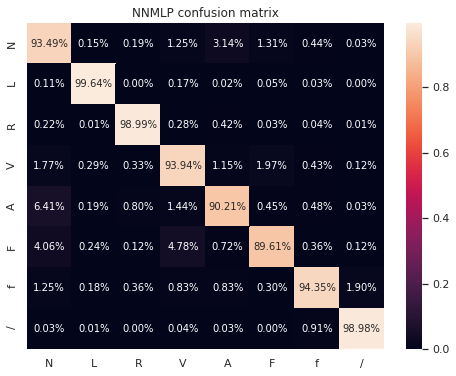}
    \includegraphics[width=0.3\textwidth, trim=0 0 60 0, clip]{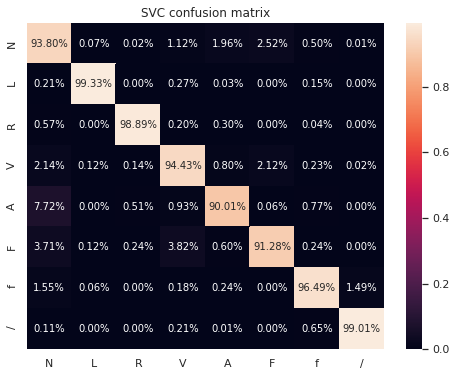}
    \includegraphics[width=0.3\textwidth, trim=0 0 60 0, clip]{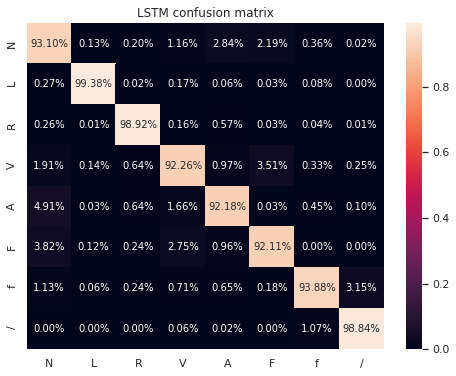}
    \includegraphics[width=0.3\textwidth, trim=0 0 60 0, clip]{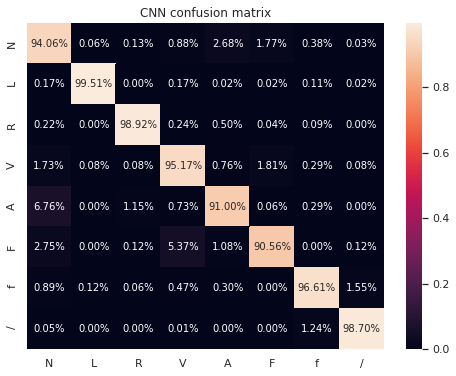}
    \caption{Examples of Single Beats of All Classes. (X-Axis: Timestamps, Y-Axis: Voltage) Cross-Validation Confusion Matrices per Model per Class. (X-Axis: Predicted Labels, Y-Axis: True Labels)}
    \label{fig:19}
\end{figure}

To visualize which how many beats were correctly classified into their respective classes and how
many were mis-classified, confusion matrices were plotted for all the models. The confusion
matrix is a symmetric 2-D matrix, where the size of the confusion matrix is the number of classes.
The higher the value of the diagonals of a confusion matrix, the higher the performance of the
classifier.

We observe in figure \ref{fig:19} that ADA and NB have the most irregular diagonals which explains their bad performance metric values, as this irregularity means the true classes do not correspond to the classified classes. As for the rest of the models, the diagonal is clearly showing 80\% and above correct classifications. We observe that for all the classifiers, L R and / classes are correctly classified with consistently over 90\%. We also notice that for N class, ADA and NB have approximately 27\% correct classifications, whereas for all other classifiers it is over 80\%. NB assumes that all features are independent, which clearly is not the case with a time-dependent ECG, hence giving poor results. As for ADA, it is extremely sensitive to Noisy data and outliers, and also learns progressively. This allows us to consider ADA and NB not fit for ECG beat classification. Analysing this by class, we see that all models give the worst result for classes A and F compared to the rest of the classes. This could be because beats of class A are very similar to class N beats. As for class F, this class has the least number of beats in the dataset and naive bootstrap up-sampling is not enough to make most models learn features accurately. Precision, recall and F1 scores for all models per class can be found github.

\subsubsection{Quantitative Analysis}

This section will analyse the performance metrics results through statistical tests like Shapiro-Wilk, Kruskal Wallis H, and Wilcoxon signed rank test \cite{RN26} \cite{RN27}. We
will investigate different quantitative analysis methods as well, like Kendall tau rank correlation
and also compare PFI and Grad-CAM values. Analysis on the cross-validation performance
metrics were implemented to obtain the 95\% Confidence Interval (CI). The Shapiro-Wilk test
tests the null hypothesis that the data was drawn from a normal distribution. The Kruskal-Wallis
H-test tests the null hypothesis that the population mean of all the groups are equal, which is a
non-parametric version of one-way Analysis of Variance (ANOVA).

\begin{minipage}{\textwidth}
  \begin{minipage}[b]{0.39\textwidth}
    \centering
    \includegraphics[width=1\textwidth]{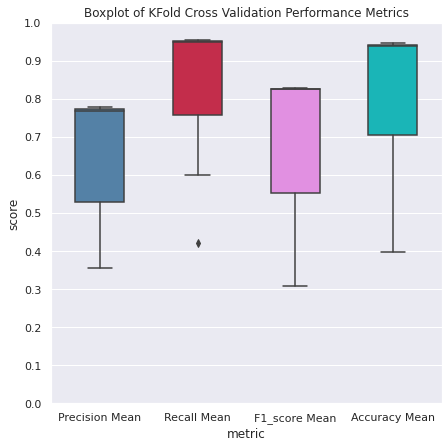}
    \captionof{figure}{All Models Performance Metrics}
  \end{minipage}
  \hfill
  \begin{minipage}[b]{0.59\textwidth}
    \centering
    \begin{center}
    \begin{tabular}{lccc}
    \hline \textbf{Variable} & \textbf{95}\% \textbf{Conf.} & \textbf{Interval} & \textbf{Kruskal} \textbf{Wallis} \textbf{H} \\
    \hline Accuracy & $0 .574$ & $1 . 0 19$ & $1.03 \mathrm{e}-5$ \\
    Precision & $0.463$ & $0 .824$ & $1.09 \mathrm{e}-5$ \\
    Recall & $0.619$ & $1.023$ & $6.78 \mathrm{e}-6$ \\
    f1 Score & $0.454$ & $0.895$ & $1.19 \mathrm{e}-5$ \\
    \hline
    \end{tabular}
    \end{center}
    \captionof{table}{Cross-Validation Statistical Analysis}
    \label{tab:4}
    \end{minipage}
  \end{minipage}

In table \ref{tab:4} the 95\% CI shows the probability of a parameter falling between values around the
mean. The confidence intervals compute the degree of certainty in a sampling method. CI of
accuracy, precision, recall, and f1 score are almost equally spread out which indicates that all of
their population means are in similar ranges. Shapiro-Wilk normality test was performed on the
metrics where most of the metrics rejected the null hypothesis but not all, hence Kruskal-Wallis
H test was performed instead of ANOVA, since non-parametric tests are more robust in this case
\cite{RN53}. The p-values of the Kruskal Wallis H test are exceedingly lower than the
significant level of 0.05, which means the null hypothesis is rejected and the population mean of
all the groups are not the same.

\begin{figure}[h]
    \centering
    \begin{subfigure}{.49\linewidth}
    \includegraphics[width=0.79\textwidth, trim=0 0 70 0, clip]{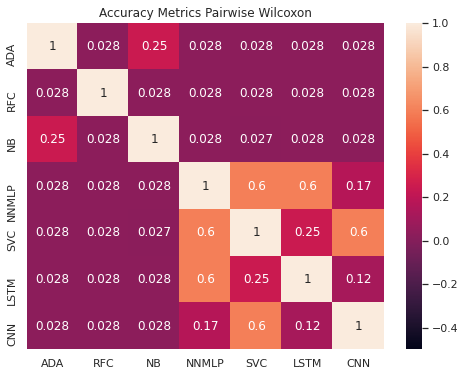}
    \caption{Accuracy Wilcoxon Test}
    \end{subfigure}
    \hfill
    \centering
    \begin{subfigure}{.49\linewidth}
    \includegraphics[width=0.79\textwidth, trim=0 0 70 0, clip]{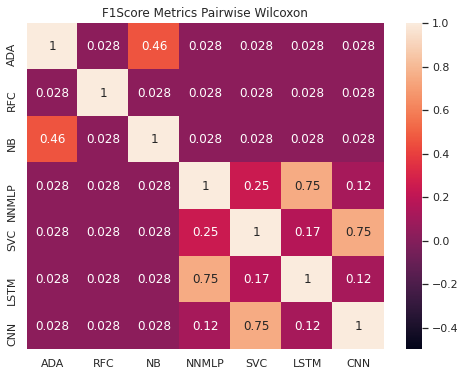}
    \caption{F1 Score Wilcoxon Test}
    \end{subfigure}
    \caption{Pairwise Wilcoxon Signed-Rank Test per Model for Performance Metrics, P-values}
    \label{fig:21}
\end{figure}

In figure \ref{fig:21}, for the Wilcoxon Signed-Rank test, the null hypothesis is rejected when that the
results are statistically significant. It is observed that pairwise comparison of NNMLP, SVC,
LSTM, and CNN have p-value greater than 0.05 which means that the null hypothesis is rejected
and the accuracy and f1 score data have identical population distribution for each cross-validation
fold. This means that the results obtained are statistically significant and not obtained by chance.

\begin{figure}[h]
    \centering
    \begin{subfigure}{.55\linewidth}
    \includegraphics[width=0.89\textwidth, trim=0 0 0 0, clip]{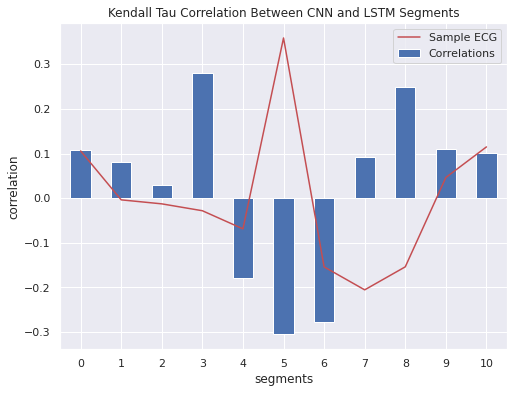}
    \caption{Per Segment Correlation between CNN and LSTM Layer Weights}
    \label{fig:221}
    \end{subfigure}
    \hfill
    \centering
    \begin{subfigure}{.44\linewidth}
    \includegraphics[width=0.85\textwidth, trim=0 0 70 0, clip]{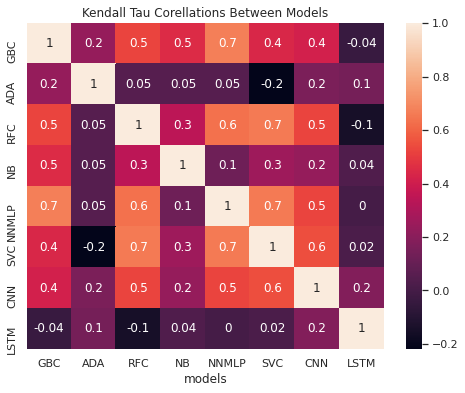}
    \caption{Per Model Correlation between Feature Weights}
    \label{fig:222}
    \end{subfigure}
    \caption{Kendall Tau Correlations P-values}
\end{figure}

Figure \ref{fig:221} shows the Kendall’s tau correlation \cite{RN36} values per segment between CNN
and LSTM layer weights. Here we observe that the values of all the segments except 4, 5, 6
are between 0.0 and 0.3 which show a moderate agreement of correlation between CNN and
LSTM weights, whereas segments 4, 5, 6 shows a moderate disagreement of correlation. This
disagreement on the QRS complex (as shown over the sample ECG) means that the layer weights
of the CNN and LSTM give different importance to the QRS complex and mostly agree on the
PR interval and the ST segment. This disagreement on the QRS complex can be visualized in
the next figure where the average Grad-CAM values for both CNN and LSTM model are not
the same at the segments 4, 5, 6. Figure \ref{fig:222} shows the pairwise Kendall’s tau correlation for the
PFI scores between models. Here it can be observed that values closer to 1 show an extremely
strong correlation while the values closer to 0 which indicates absence of association through null hypothesis. PFI values of ADA, NB and LSTM have the almost no correlation with other models.
This could be due to the bad performance of ADA and NB models, whereas for the LSTM model,
it could be the results of the feature importance scores favouring completely different segments
than the other models. This can be better seen in the next figure which visualizes the PFI values
for the LSTM model.

\begin{figure}[h]
    \centering
    \begin{subfigure}{.49\linewidth}
    \includegraphics[width=0.99\textwidth, trim=0 0 0 0, clip]{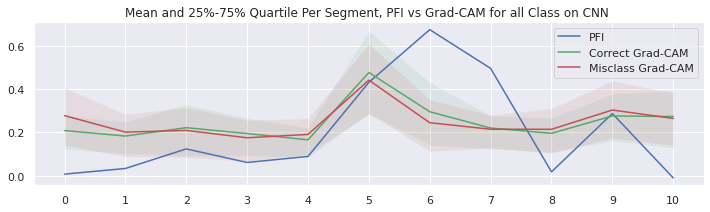}
    \includegraphics[width=0.99\textwidth, trim=0 0 0 0, clip]{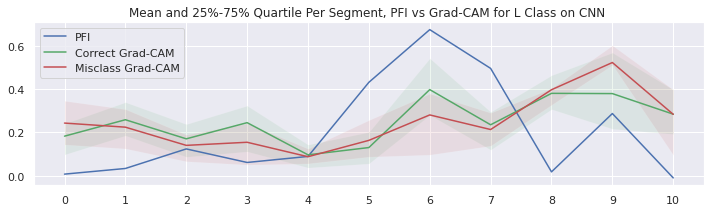}
    \includegraphics[width=0.99\textwidth, trim=0 0 0 0, clip]{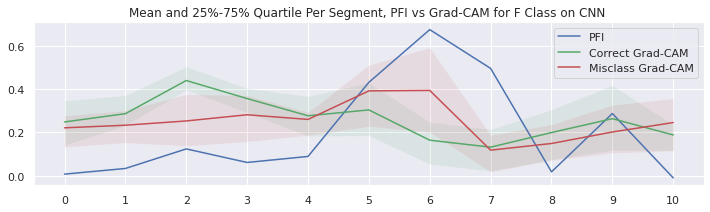}
    \caption{CNN Model}
    \end{subfigure}
    \hfill
    \begin{subfigure}{.49\linewidth}
    \includegraphics[width=0.99\textwidth, trim=0 0 0 0, clip]{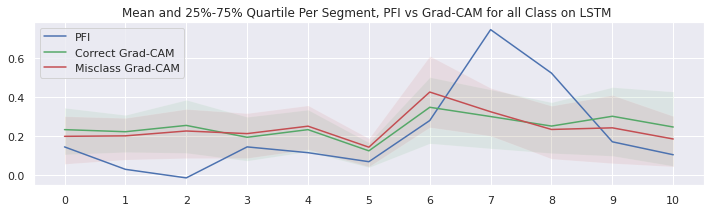}
    \includegraphics[width=0.99\textwidth, trim=0 0 0 0, clip]{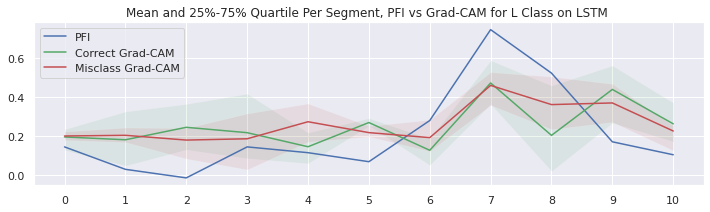}
    \includegraphics[width=0.99\textwidth, trim=0 0 0 0, clip]{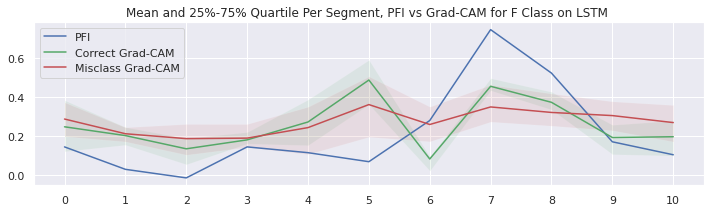}
    \caption{LSTM Model}
    \end{subfigure}
    \caption{Mean and 25\%-75\% Quantile Values Per Segment - PFI vs Grad-CAM Average (top), L Class (middle), F Class (bottom), for CNN and LSTM Models. (X-Axis: Segments, Y-Axis: Scaled Values)}
    \label{fig:23}
\end{figure}

We then move on to quantitatively analyse and compare PFI feature importance weights and
Grad-CAM correct and mis-classified results per segment. This is done by scaling the results and
plotting the mean and 25\% - 75\% quantile range of Grad-CAM values along with the obtained
PFI values. As seen in figure \ref{fig:23} we observe that the average plot for CNN and LSTM focus on
the QRS complex and the ST segment, respectively. For the L class plots, PFI and Grad-CAM
values agree with each other as they both focus on the same segments. As for F class plots, the
values did not quite agree with the important segments. This is because F class is more likely to
be mis-classified than the L class, as seen in the confusion matrices results.

\subsection{Leave Groups Out (Holdout Patients)}

After K-fold validation on traditional holdout method which tested on unseen beats, the Leave
Groups Out approach was used to evaluate the classifier performance. This method allows us
to test on unseen patients as patient number 104, 113, 119, 208, and 210 were split from the
dataset to be used for testing, while the remaining 43 patients were used for training. Holding
out patients from the dataset instead of beats leads to the risk of also removing classes of ECG
beats, since a few abnormalities may be limited to one or two patients. The same method has also
been used by \cite{RN32}, where they also chose the same patient numbers to hold out.
The selected patients however do not contain any beats from classes L, R, and A.

\begin{figure}[h]
    \centering
    \includegraphics[width=0.8\textwidth]{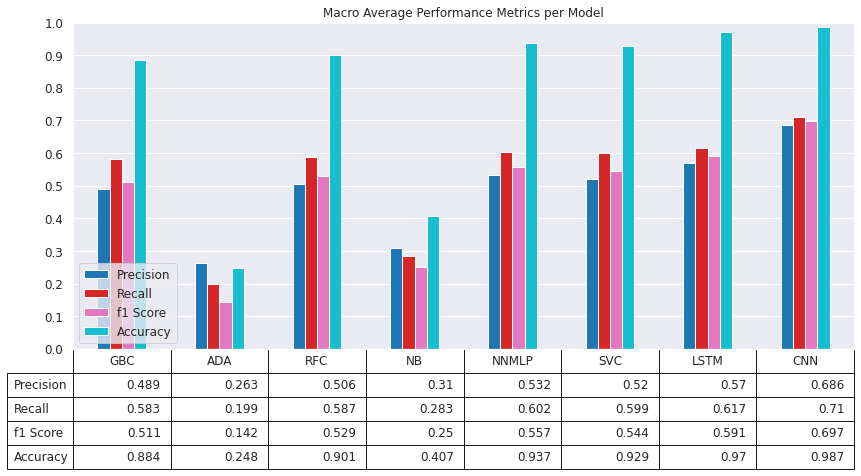}
    \caption{Macro Average Leave Groups Out Performance Metrics per Model. (X-Axis: Models, Y-Axis: Scores)}
    \label{fig:24}
\end{figure}

\begin{figure}[h]
    \centering
    \includegraphics[width=0.3\textwidth, trim=0 0 60 0, clip]{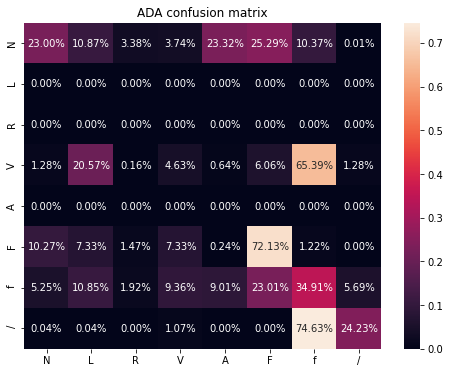}
    \includegraphics[width=0.3\textwidth, trim=0 0 60 0, clip]{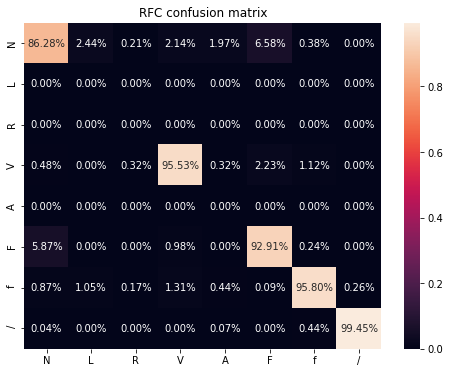}
    \includegraphics[width=0.3\textwidth, trim=0 0 60 0, clip]{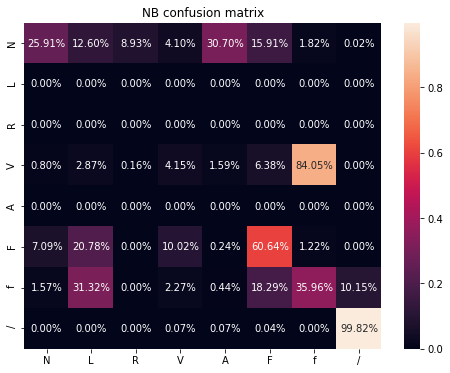}
    \includegraphics[width=0.3\textwidth, trim=0 0 60 0, clip]{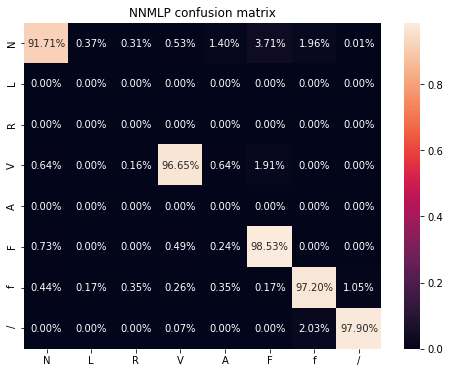}
    \includegraphics[width=0.3\textwidth, trim=0 0 60 0, clip]{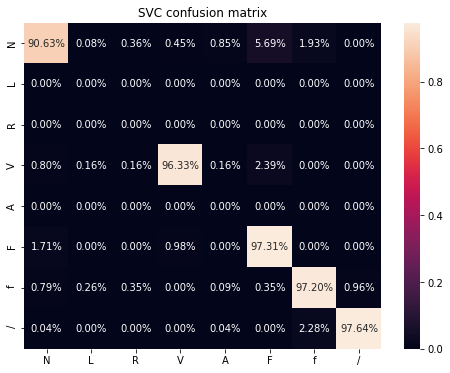}
    \includegraphics[width=0.3\textwidth, trim=0 0 60 0, clip]{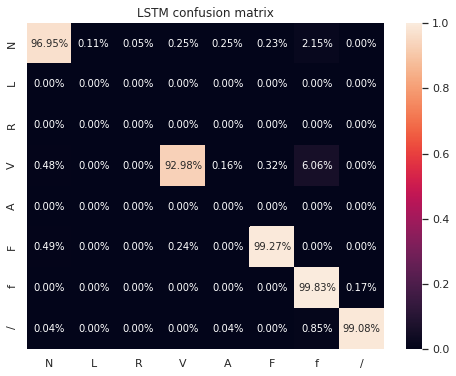}
    \includegraphics[width=0.3\textwidth, trim=0 0 60 0, clip]{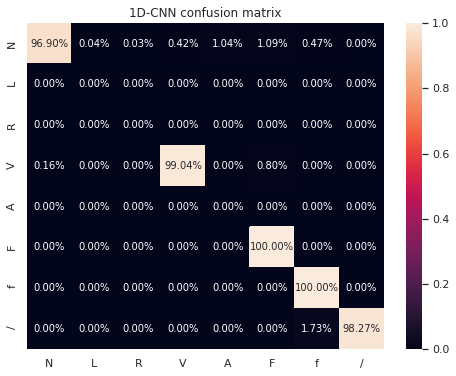}
    \caption{Examples of Single Beats of All Classes. (X-Axis: Timestamps, Y-Axis: Voltage) Cross-Validation Confusion Matrices per Model per Class. (X-Axis: Predicted Labels, Y-Axis: True Labels)}
    \label{fig:25}
\end{figure}

Patient holdout method allowed us to test on beats from unseen patients, whereas the beat holdout
method shuffled beats of all patients for training and testing purposes. Although testing using
the beat holdout method is considered the most common method, it can result in flawed results
due to data leakage. Separating ECG recordings by beats and randomly splitting a shuffled set of
these beats for testing results in having beats in both the training and testing dataset from the same patient. Due to similar characteristics of patients like electrode placement, pre-existing
conditions, and medications, there is a chance of common morphology between beats. In simpler
terms, a beat from one patient will probably look same as another beat (of the same class) from
the same patient. This results in data bleeding between the training and dataset, which should be
avoided \cite{RN32}.

\subsubsection{Classification Performance Metrics}

Figure \ref{fig:24} shows the macro average performance metric results per class for the patient holdout
method. It also shows results for GBC, which was discarded for K-Fold cross-validation. We see
that it follows the same trend as before in terms of accuracy, where CNN and LSTM are the best
performing models with both of them having accuracy of more than 97\%. The precision, recall
and f1 score metric give poor results for the patient holdout method where the NNMLP, SVC,
CNN, and LSTM scores drop to between 55\% and 70\%, compared to beats holdout method.
The main reason for this is class imbalance since leaving patients out from training dataset is
resulting in limited number of beats from one class to be present hence the lower precision and
recall. To see scores corrected of this imbalance, the weighted average results can be found on github, where the proportions of the labels of each class are considered.

The low score could also be an issue of unseen data but looking at the confusion matrices in figure
\ref{fig:25}, it is clear that CNN and LSTM models do not struggle with this issue, giving accuracy of
92\% and above for all classes present in the test dataset. It is observed that classes L, R, and A are
missing from the test data, which is one of the disadvantages of using Leave Groups Out method.
Similarly as before, the diagonals for ADA and NB are irregular indicating bad performance. As for the CNN model which is the best performing model,
it obtains 100\% accuracy for classifying the F and f class, this is happening due to the unique
morphology of fusion of paced beats as they are not completely natural and are partially created
by a pacemaker. This allows us to conclude that the model performs well on unseen data as well.

\subsubsection{Quantitative Analysis}

\begin{minipage}{\textwidth}
  \begin{minipage}[b]{0.49\textwidth}
    \centering
    \includegraphics[width=1\textwidth]{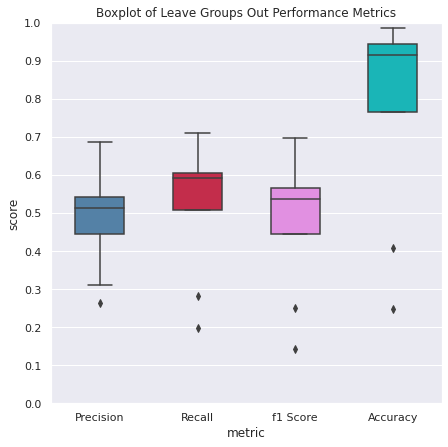}
    \captionof{figure}{All Models Performance Metrics}
  \end{minipage}
  \hfill
  \begin{minipage}[b]{0.49\textwidth}
    \centering
    \begin{center}
    \begin{tabular}{lcc}
    \hline \textbf{Variable} & \textbf{95}\% \textbf{Conf.} & \textbf{Interval} \\
    \hline Accuracy & $0.543$ & $1.022$ \\
    Precision & $0.369$ & $0.599$ \\
    Recall & $0.372$ & $0.672$ \\
    f1 Score & $0.322$ & $0.632$ \\
    \hline
    \end{tabular}
    \end{center}
    \captionof{table}{Leave Groups Out Statistical Analysis}
    \label{tab:5}
    \end{minipage}
  \end{minipage}

In the analysis of Leave Groups Out method, tests like Shapiro-Wilk, Kruskal-Wallis H, and
Wilcoxon Signed Rank were not implemented since there were no multiple folds to statistically
test the data on. However, the rest of the evaluation of the models follow the same pattern as the
previous section with the implementation 95\% CI on performance metrics and the Kendall tau
rank correlation on PFI and Grad-CAM values. Table \ref{tab:5} shows that the CI of precision, recall
and f1 score are in the same range, while the CI for accuracy is more spaced out, indicating that
the population means of accuracy is in a larger range than the other metrics.

\begin{figure}[h]
    \centering
    \begin{subfigure}{.55\linewidth}
    \includegraphics[width=0.89\textwidth, trim=0 0 0 0, clip]{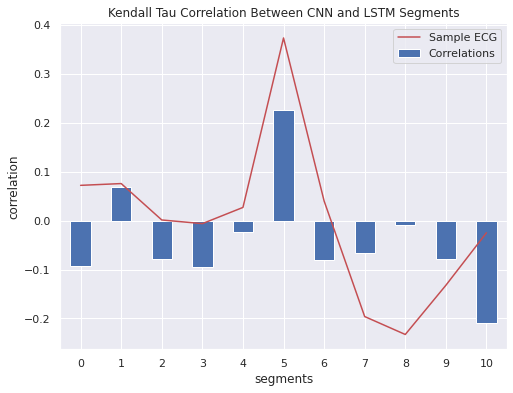}
    \caption{Per Segment Correlation between CNN and LSTM Layer Weights}
    \label{fig:271}
    \end{subfigure}
    \hfill
    \centering
    \begin{subfigure}{.44\linewidth}
    \includegraphics[width=0.85\textwidth, trim=0 0 70 0, clip]{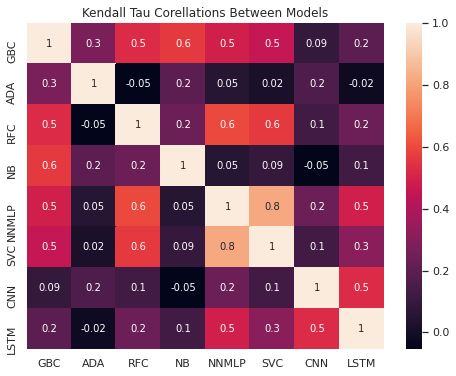}
    \caption{Per Model Correlation between Feature Weights}
    \label{fig:272}
    \end{subfigure}
    \caption{Kendall Tau Correlations P-values}
    
\end{figure}

Figure \ref{fig:271} shows the Kendall’s tau correlation values per segment between the LSTM and
CNN layer weights. Here it is observed that segment 5 shows moderate agreement, and segment
10 shows moderate disagreement of correlation, and the rest of the segments are showing no
monotonous relation at all. This agreement in the middle of the QRS complex can be visualized
in the next figure where the average Grad-CAM values of both CNN and LSTM model are
same at the 5th segment. Figure \ref{fig:272} shows the pairwise Kendall’s tau correlation for the PFI
scores between models. PFI values of ADA, NB and CNN have the almost no correlation with
other models. This could be due to the bad performance of the ADA and NB models, whereas for
CNN model, it could be a result of the feature importance scores favouring completely different
segments than the other models. This can be better seen in the next figure which visualizes the
PFI values for the CNN model.

\begin{figure}[h]
    \centering
    \begin{subfigure}{.49\linewidth}
    \includegraphics[width=0.99\textwidth, trim=0 20 0 0, clip]{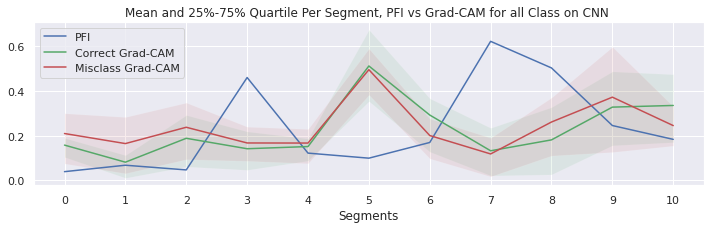}
    \includegraphics[width=0.99\textwidth, trim=0 20 0 0, clip]{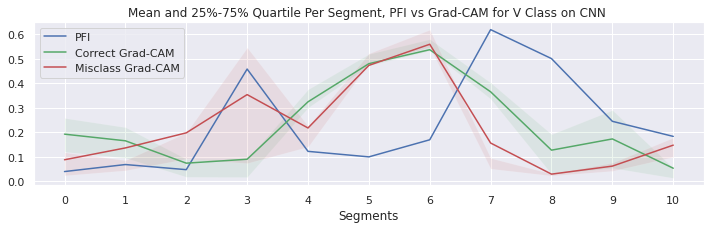}
    \includegraphics[width=0.99\textwidth, trim=0 20 0 0, clip]{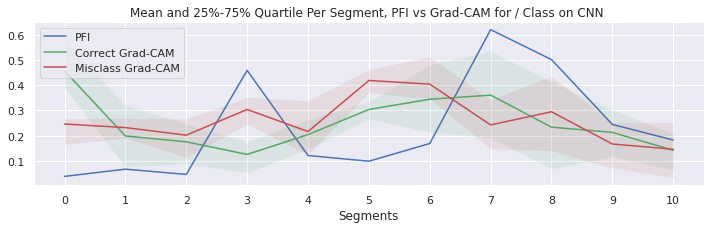}
    \caption{CNN Model}
    \end{subfigure}
    \hfill
    \begin{subfigure}{.49\linewidth}
    \includegraphics[width=0.99\textwidth, trim=0 0 0 0, clip]{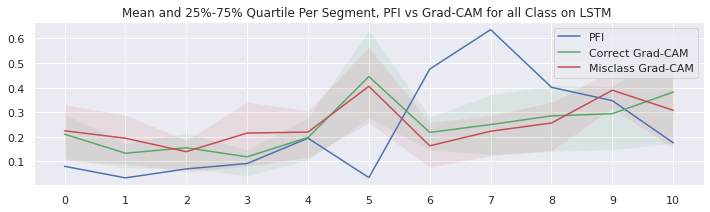}
    \includegraphics[width=0.99\textwidth, trim=0 0 0 0, clip]{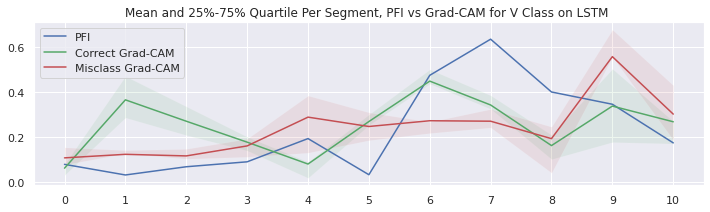}
    \includegraphics[width=0.99\textwidth, trim=0 0 0 0, clip]{images/metrics/leave/LSTM_V_pfigrad.png}
    \caption{LSTM Model}
    \end{subfigure}
    \caption{Mean and 25\%-75\% Quantile Values Per Segment - PFI vs Grad-CAM Average (top), V Class (middle), / Class (bottom), for CNN and LSTM Models. (X-Axis: Segments, Y-Axis: Scaled
Values)}
    \label{fig:28}
\end{figure}

We then move on to quantitatively analyse and compare PFI feature importance weights and
Grad-CAM correct and mis-classified results per segment. The same L and F classes as the
previous section are not compared in this section, as the CNN model predicted the F class with
100\% accuracy leaving no incorrect beats to compare, and the patient holdout method did not
contain the L class beats. In figure \ref{fig:28} we observe that average CNN and LSTM Grad-CAM
values focus on segment 5 of the QRS complex, while the PFI values for both the models lean
towards segments 7, 8, 9 i.e. the ST segment. Along with looking at the ST segment, the PFI
for CNN model also focuses on the PR interval giving a bump at segment 3. For class V, the
CNN model has high Grad-CAM values on the QRS complex, while the LSTM model considers
segments 1 and 6 more important. For class /, the CNN model seems to be slightly considering
segments 6 and 7 as important, while the LSTM model focuses heavily on segments 5, 6, 7, and
9 i.e., the later part of the QRS complex and ST segment. These inconsistencies that are found in
the patient holdout interpretability were not present in the traditional beats holdout method.

\subsection{Variance per Segment}

\begin{figure}[h]
    \centering
    \begin{subfigure}{.49\linewidth}
    \includegraphics[width=0.99\textwidth, trim=0 0 0 0, clip]{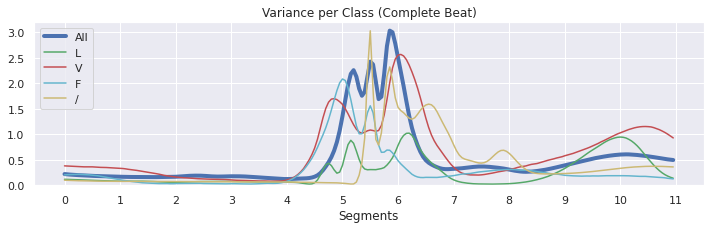}
    \caption{Complete Signal}
    \end{subfigure}
    \hfill
    \centering
    \begin{subfigure}{.49\linewidth}
    \includegraphics[width=0.99\textwidth, trim=0 0 0 0, clip]{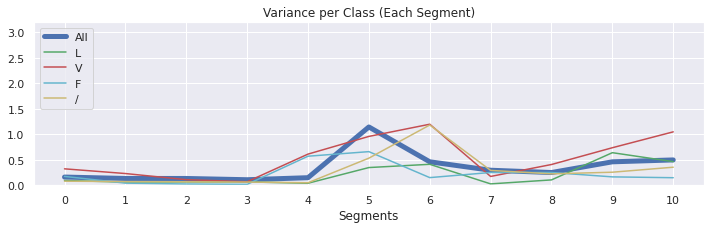}
    \caption{Mean per Segment}
    \end{subfigure}
    \caption{Variance per Segment for Classes L, F, f, /, and All Beats. (X-Axis: Segments, Y-Axis:
Scaled Values)}
    \label{fig:29}
\end{figure}

To better understand the Grad-CAM vs PFI comparison plots, variance of segments of the ECG
beats per class was plotted as seen in figure \ref{fig:29}. We are looking at the variance of each segment
to understand where in the signal the classes differ the most, and it would be expected that the
segments with the highest variance would be picked up by the ML models, and therefore tells
the algorithm most about which class each beat belongs to. It is seen that figure 4.12 has classes
L, F, V and / which were previously evaluated in various sections. Class L contains more variance in the ST segment compared to the QRS complex,
whereas class F contains variance mostly in the QRS complex. As for /, high variance is observed
in the QRS complex, whereas class V shows high variance in both QRS complex and ST segment.
The thick blue line shows variance across all beats which has high values concentrated to the
middle of the QRS complex. Variance per segment plots for all classes can be found in on github. Taking these variances into account, it is observed that the plots for Grad-CAM
comparison per class in figures \ref{fig:23} and \ref{fig:28} show consistency with the model’s predictions on
the segments with the most amount of variance.

\section{Discussion and Conclusion}

\subsection{Summary}

In this study, different interpretability tools to explain ML models for time-series ECG clas-
sification problem were investigated. Firstly, the MIT-BIH arrhythmia dataset was split into
ECG beats of eight annotation classes, and then ECG classification was performed using eight
different classifiers. Next, the ECG beats were divided into 11 segments which would allow us to
understand which area of the beats per class were focused on by the models during the prediction
making process. K-fold cross-validation (holdout beats) and Leave Groups Out (holdout patients)
were the two model evaluation techniques used on the classifiers. Four different interpretability
techniques were applied on the models and the resulting explanations were investigated and
evaluated using quantitative statistical analysis. The following section will discuss the investigated interpretability techniques before laying down the conclusion and talking about limitations and future work.

\subsection{Discussion on Interpretability Techniques}

One of the biggest disadvantages of PDP is that there is an assumption of independence i.e., it is
assumed that features used for computing the partial dependence are not correlated between each
other. However, if the features are not correlated, then the PDP display the average influence of
the feature in the predictions. The main issue is also that the heterogeneous effects are hidden in
PDP as it only displays the marginal effect each class or feature has in the model’s decision-making
process. The easy to implement nature of PDP is not advantageous enough to outweigh the
issues it faces for time-series interpretability. In this case, the clarity of explanation provided
by PDP were uncertain and the metrics were unavailable, which allows us to conclude that this
method is not fit for explaining models classifying time-series data.

KernelSHAP is extremely slow which makes it difficult to implement it for many instances.
In this case, there are more than 30,000 beats to be tested but only the first 50 instances are
used to produce the decision and summary plots for KernelSHAP. For time-series, SHAP has a
possibility of hiding biases and intentionally displaying misleading interpretation since it is only
designed to deal with 2D data. The only advantage is that with SHAP, global interpretations
are consistent with the local explanations. For my investigation, KernelSHAP did provide
satisfying explanations which were consistent with both global and local interpretations of the
SVC model, which allowed us to look at the overall decision for the dataset along with explaining
the behaviour of the model for each class. However, the timing overhead, incompatibility with
deep and recurrent neural networks, and the small number of instances explained by this method
makes it clear that SHAP is not fit for explaining models classifying time-series data.

Permutation feature importance is the increase in model error when the feature’s information
is destroyed, which provides an extremely compressed, global interpretation into the model’s
decision-making behaviour. It has a few advantages which include no model retraining, accounting for all interactions between features, and it is comparable across all data types like tabular,
text, time-series etc. One of the disadvantages is that it depends on reshuffling features, adding randomness to the data measurements. The model agnostic and global interpretability nature
of this method proves to be fairly useful for explaining the model’s behaviour as the results are
accurate and consist. However, unlike SHAP there is no local interpretability, which means it is
not possible to explain the behaviour of ECG segments for each class of this data.

Grad-CAM created saliency maps which showed each data point’s unique quality per beat which
allowed for a very local explanation of the model’s predictive process. These produced coarse
localization maps of the important regions allowing us to visualize which segments are more
heavily influencing the model’s behaviour. We then approximated global interpretability by
averaging the local maps per class. These are implemented on the output of the final 1D-CNN or
LSTM layers. This implementation is limited to TensorFlow models making it a model specific
interpretability method. However, having ability to both locally and globally explain deep and
recurrent neural network models makes this implementation a very versatile interpretability
technique which also allows the users to understand the explanations, unlike PDP. For this reason,
Grad-CAM satisfied all the mentioned attributes of interpretability and fidelity of the model’s
explanation.

For each explainable interpretability method, it was designated if a attribute is satisfied (S), or if
an attribute is unsatisfied (U). All the interpretability methods implemented in this investigation
were post-hoc explanations of the model’s decisions, which means the interpretation occurred
after the models were finished running. These post-hoc explanations can be seen in the table \ref{tab:6}.

To reach a set of pragmatic definitions, interpretability and fidelity were split into general
fundamental factors that in conjunction determine the quality of an explanation. This method of
qualitative evaluation of interpretability techniques is proposed by \cite{RN38}. This leads
to the following definitions. Interpretability: An explanation is interpretable if it is unambiguous
and yields a single reasoning that is similar for related instances (clarity), the explanation is not
too complicated and is presented in a condensed form (parsimony). Interpretability describes
the degree to which a user can understand an explanation. Fidelity: An explanation is accurate
if the explanation describes the complete dynamic of the given model and provides adequate
information to calculate the output for a given input (completeness) and the explanation is correct
and is candid to the model’s task (soundness). Fidelity describes the graphic accuracy of the
explanation.

\begin{table}[h]
    \centering
    \caption{Evaluating Post-hoc Interpretability Methods}
    \begin{tabular}{l|cccc|cc}
    \hline \textbf{Methods} & \multicolumn{4}{c|}{ \textbf{Interpretability} } & \multicolumn{2}{c}{ \textbf{Fidelity} } \\
    \hline
    & Scope & Model & Clarity & Parsimony & Completeness & Soundness \\
    \hline PDP & Global & Specific & U & S & U & $U$ \\
    SHAP & Local & Specific & S & S & U & S \\
    PFI & Global & Agnostic & S & S & U & S \\
    Grad-CAM & Local & Specific & S & S & S & S \\
    \hline
    \end{tabular}
    \label{tab:6}
\end{table}

\subsection{Conclusion}

The investigation of model interpretability is vital for the application of these models for clinical
usage in the healthcare industry. The motivation of this project was to find the best performing
interpretability technique to explain various models’ prediction process for time-series ECG
classification task.

The results of the ECG classification task showed that the convolutional neural network and
the long short-term memory classifiers were the best performing models for both K-Fold cross-
validation and Leave Groups Out model evaluation methods. The obtained accuracy for CNN and LSTM models were: 94.1\% and 94\% for K-Fold cross-validation method, and 98.7\% and 97\%
for Leave Groups Out method. Quantitative hypothesis testing backed the resulting performance
metrics in terms of significance levels, which allowed us to conclude that the classification of
CNN and LSTM were consistently better than other models. These two neural networks also
had the advantage of being interpretable by both Grad-CAM and PFI, which provided both
local and global explanations of their prediction process.

In my conclusion of the investigation of interpretability methods, saliency maps using Grad-CAM
proved to be the most effective technique to locally interpret neural networks. The advantage
of averaging Grad-CAM and neural network layer weights per class and also for the whole
dataset, allowed the explanation of the model’s predictions both locally and globally for individual
data points, and segments of the ECG beats. The versatile nature of implementing Grad-CAM
interpretability on time-series classification made it the most capable technique compared to
the rest. PFI had the advantage of being the only model agnostic interpretability technique
which explained TensorFlow classifier’s layer weights using a surrogate model, but gave only
global explanations of the segments for the entire dataset, and not for each class. Quantitatively
comparing PFI and Grad-CAM values allowed for deeper analysis of the difference between local
and global explanations. PDP was found to be extremely uninformative at explaining different
segments of the ECG beat, while SHAP was too time consuming and explained a very limited
amount of data. This allowed me to conclude that PDP and SHAP methods were the worst
performing interpretability techniques for ECG data.

Investigating these techniques allowed me to conclude that the proposed neural network and
SK-Learn classifiers can be successfully interpreted for time-series ECG classification and are not
complete black boxes. The model’s ability to accurately classify the ECG beats by considering
the QRS complex more significant than other ECG segments, proved that the prediction process
agrees with medical literature i.e., how clinicians and cardiologists classify abnormal ECG beats.

\subsection{Limitations and Future Work}

There is a significant imbalance of classes of ECG beats in the MIT-BIH
dataset which makes the model struggle to learn differences between classes even after resampling.
The beats are taken from a small number of patients which limits diversity in characteristics
of heartbeats. Secondly, most interpretability software and libraries are not created to explain
time-series data due to abundance of data points in one signal. These methods explain the
model through features of the data and providing 220 data points each for 32,000 beats would
not be feasible. Hence, the ECG beats needed to be segmented and given as features to the
interpretability software which made it lose significant amount of data points. As for Grad-CAM,
one of the limitations is that it struggles to localize multiple occurrences of the same features
and may provide false localization of heat-maps with reference to class region. It also gives no
explanation of interaction between features and combinations of inputs.

Future work can involve testing larger arrhythmia datasets on better and more fine-tuned
deep learning models. Alternative to Grad-CAM attribution method, integrated gradients
\cite{RN49} could be used to investigate model interpretability on time-series ECG
data for more accurate localization of features. The investigation of attention mechanisms \cite{RN30} to recognize critically significant ECG segments associated to model performance
could lead to better interpretability, compared to using gradient weights to explain the model’s
prediction process.

\bibliographystyle{unsrtnat}
\bibliography{references}

\begin{thebibliography}{28}
\providecommand{\natexlab}[1]{#1}
\providecommand{\url}[1]{\texttt{#1}}
\expandafter\ifx\csname urlstyle\endcsname\relax
  \providecommand{\doi}[1]{doi: #1}\else
  \providecommand{\doi}{doi: \begingroup \urlstyle{rm}\Url}\fi

\bibitem[Wu et~al.(2018)Wu, Gale, Hall, Dondo, Metcalfe, Oliver, Batin,
  Hemingway, Timmis, and West]{RN54}
Jianhua Wu, Chris~P Gale, Marlous Hall, Tatendashe~B Dondo, Elizabeth Metcalfe,
  Ged Oliver, Phil~D Batin, Harry Hemingway, Adam Timmis, and Robert~M West.
\newblock Editor’s choice-impact of initial hospital diagnosis on mortality
  for acute myocardial infarction: A national cohort study.
\newblock \emph{European Heart Journal: Acute Cardiovascular Care}, 7\penalty0
  (2):\penalty0 139--148, 2018.
\newblock ISSN 2048-8726.

\bibitem[Moody and Mark(2001)]{RN41}
George~B Moody and Roger~G Mark.
\newblock The impact of the mit-bih arrhythmia database.
\newblock \emph{IEEE Engineering in Medicine and Biology Magazine}, 20\penalty0
  (3):\penalty0 45--50, 2001.
\newblock ISSN 0739-5175.

\bibitem[Molnar(2019)]{RN40}
C~Molnar.
\newblock Interpretable machine learning. leanpub.
\newblock \emph{Victoria}, 2019.

\bibitem[Lundberg and Lee(2017)]{RN37}
Scott~M Lundberg and Su-In Lee.
\newblock A unified approach to interpreting model predictions.
\newblock \emph{Advances in neural information processing systems}, 30, 2017.

\bibitem[Altmann et~al.(2010)Altmann, Toloşi, Sander, and Lengauer]{RN25}
André Altmann, Laura Toloşi, Oliver Sander, and Thomas Lengauer.
\newblock Permutation importance: a corrected feature importance measure.
\newblock \emph{Bioinformatics}, 26\penalty0 (10):\penalty0 1340--1347, 2010.
\newblock ISSN 1460-2059.

\bibitem[Selvaraju et~al.(2017)Selvaraju, Cogswell, Das, Vedantam, Parikh, and
  Batra]{RN47}
Ramprasaath~R Selvaraju, Michael Cogswell, Abhishek Das, Ramakrishna Vedantam,
  Devi Parikh, and Dhruv Batra.
\newblock Grad-cam: Visual explanations from deep networks via gradient-based
  localization.
\newblock In \emph{Proceedings of the IEEE international conference on computer
  vision}, pages 618--626, 2017.

\bibitem[Velayudhan and Peter(2016)]{RN51}
Aswathy Velayudhan and Soniya Peter.
\newblock Noise analysis and different denoising techniques of ecg signal-a
  survey.
\newblock \emph{IOSR journal of electronics and communication engineering},
  1\penalty0 (1):\penalty0 40--44, 2016.

\bibitem[Kelly et~al.(2019)Kelly, Karthikesalingam, Suleyman, Corrado, and
  King]{RN34}
Christopher~J Kelly, Alan Karthikesalingam, Mustafa Suleyman, Greg Corrado, and
  Dominic King.
\newblock Key challenges for delivering clinical impact with artificial
  intelligence.
\newblock \emph{BMC medicine}, 17\penalty0 (1):\penalty0 1--9, 2019.
\newblock ISSN 1741-7015.

\bibitem[Kachuee et~al.(2018)Kachuee, Fazeli, and Sarrafzadeh]{RN33}
Mohammad Kachuee, Shayan Fazeli, and Majid Sarrafzadeh.
\newblock Ecg heartbeat classification: A deep transferable representation.
\newblock In \emph{2018 IEEE international conference on healthcare informatics
  (ICHI)}, pages 443--444. IEEE, 2018.
\newblock ISBN 153865377X.

\bibitem[Acharya et~al.(2017)Acharya, Fujita, Lih, Hagiwara, Tan, and
  Adam]{RN24}
U~Rajendra Acharya, Hamido Fujita, Oh~Shu Lih, Yuki Hagiwara, Jen~Hong Tan, and
  Muhammad Adam.
\newblock Automated detection of arrhythmias using different intervals of
  tachycardia ecg segments with convolutional neural network.
\newblock \emph{Information sciences}, 405:\penalty0 81--90, 2017.
\newblock ISSN 0020-0255.

\bibitem[Pandey and Janghel(2019)]{RN46}
Saroj~Kumar Pandey and Rekh~Ram Janghel.
\newblock Automatic detection of arrhythmia from imbalanced ecg database using
  cnn model with smote.
\newblock \emph{Australasian physical and engineering sciences in medicine},
  42\penalty0 (4):\penalty0 1129--1139, 2019.
\newblock ISSN 1879-5447.

\bibitem[Gao et~al.(2019)Gao, Zhang, Lu, and Wang]{RN28}
Junli Gao, Hongpo Zhang, Peng Lu, and Zongmin Wang.
\newblock An effective lstm recurrent network to detect arrhythmia on
  imbalanced ecg dataset.
\newblock \emph{Journal of healthcare engineering}, 2019, 2019.
\newblock ISSN 2040-2295.

\bibitem[Mousavi and Afghah(2019)]{RN42}
Sajad Mousavi and Fatemeh Afghah.
\newblock Inter-and intra-patient ecg heartbeat classification for arrhythmia
  detection: a sequence to sequence deep learning approach.
\newblock In \emph{ICASSP 2019-2019 IEEE International Conference on Acoustics,
  Speech and Signal Processing (ICASSP)}, pages 1308--1312. IEEE, 2019.
\newblock ISBN 1479981311.

\bibitem[Stiglic et~al.(2019)Stiglic, Kocbek, Fijacko, Zitnik, Verbert, and
  Cilar]{RN48}
G~Stiglic, P~Kocbek, N~Fijacko, M~Zitnik, K~Verbert, and L~Cilar.
\newblock Interpretability of machine learning based prediction models in
  healthcare. arxiv 2020.
\newblock \emph{arXiv preprint arXiv:2002.08596}, 2019.

\bibitem[Vijayarangan et~al.(2020)Vijayarangan, Murugesan, Vignesh, Preejith,
  Joseph, and Sivaprakasam]{RN52}
Sricharan Vijayarangan, Balamurali Murugesan, R~Vignesh, SP~Preejith, Jayaraj
  Joseph, and Mohansankar Sivaprakasam.
\newblock Interpreting deep neural networks for single-lead ecg arrhythmia
  classification.
\newblock In \emph{2020 42nd Annual International Conference of the IEEE
  Engineering in Medicine and Biology Society (EMBC)}, pages 300--303. IEEE,
  2020.
\newblock ISBN 1728119901.

\bibitem[Jones et~al.(2020)Jones, Deligianni, and Dalton]{RN32}
Yola Jones, Fani Deligianni, and Jeff Dalton.
\newblock Improving ecg classification interpretability using saliency maps.
\newblock In \emph{2020 IEEE 20th International Conference on Bioinformatics
  and Bioengineering (BIBE)}, pages 675--682. IEEE, 2020.
\newblock ISBN 1728195748.

\bibitem[Pedregosa et~al.(2011)Pedregosa, Varoquaux, Gramfort, Michel, Thirion,
  Grisel, Blondel, Prettenhofer, Weiss, and Dubourg]{RN44}
Fabian Pedregosa, Gaël Varoquaux, Alexandre Gramfort, Vincent Michel, Bertrand
  Thirion, Olivier Grisel, Mathieu Blondel, Peter Prettenhofer, Ron Weiss, and
  Vincent Dubourg.
\newblock Scikit-learn: Machine learning in python.
\newblock \emph{the Journal of machine Learning research}, 12:\penalty0
  2825--2830, 2011.
\newblock ISSN 1532-4435.

\bibitem[Jolliffe(2011)]{RN31}
Ian Jolliffe.
\newblock Principal component analysis (pp. 1094-1096).
\newblock \emph{Springer Berlin Heidelberg. RESUME SELİN DEĞİRMECİ Marmara
  University, Goztepe Campus ProQuest Number: ProQuest). Copyright of the
  Dissertation is held by the Author. All Rights Reserved}, 28243034:\penalty0
  28243034, 2011.

\bibitem[Van~der Maaten and Hinton(2008)]{RN50}
Laurens Van~der Maaten and Geoffrey Hinton.
\newblock Visualizing data using t-sne.
\newblock \emph{Journal of machine learning research}, 9\penalty0 (11), 2008.
\newblock ISSN 1532-4435.

\bibitem[McInnes et~al.(2018)McInnes, Healy, and Melville]{RN39}
Leland McInnes, John Healy, and James Melville.
\newblock Umap: Uniform manifold approximation and projection for dimension
  reduction.
\newblock \emph{arXiv preprint arXiv:1802.03426}, 2018.

\bibitem[Kher(2019)]{RN35}
Rahul Kher.
\newblock Signal processing techniques for removing noise from ecg signals.
\newblock \emph{J. Biomed. Eng. Res}, 3\penalty0 (101):\penalty0 1--9, 2019.

\bibitem[Carrasco et~al.(2020)Carrasco, García, Rueda, Das, and Herrera]{RN26}
Jacinto Carrasco, Salvador García, MM~Rueda, Swagatam Das, and Francisco
  Herrera.
\newblock Recent trends in the use of statistical tests for comparing swarm and
  evolutionary computing algorithms: Practical guidelines and a critical
  review.
\newblock \emph{Swarm and Evolutionary Computation}, 54:\penalty0 100665, 2020.
\newblock ISSN 2210-6502.

\bibitem[Demšar(2006)]{RN27}
Janez Demšar.
\newblock Statistical comparisons of classifiers over multiple data sets.
\newblock \emph{The Journal of Machine Learning Research}, 7:\penalty0 1--30,
  2006.
\newblock ISSN 1532-4435.

\bibitem[Virtanen et~al.(2020)Virtanen, Gommers, Oliphant, Haberland, Reddy,
  Cournapeau, Burovski, Peterson, Weckesser, and Bright]{RN53}
Pauli Virtanen, Ralf Gommers, Travis~E Oliphant, Matt Haberland, Tyler Reddy,
  David Cournapeau, Evgeni Burovski, Pearu Peterson, Warren Weckesser, and
  Jonathan Bright.
\newblock Scipy 1.0: fundamental algorithms for scientific computing in python.
\newblock \emph{Nature methods}, 17\penalty0 (3):\penalty0 261--272, 2020.
\newblock ISSN 1548-7105.

\bibitem[Knight(1966)]{RN36}
William~R Knight.
\newblock A computer method for calculating kendall's tau with ungrouped data.
\newblock \emph{Journal of the American Statistical Association}, 61\penalty0
  (314):\penalty0 436--439, 1966.
\newblock ISSN 0162-1459.

\bibitem[Markus et~al.(2021)Markus, Kors, and Rijnbeek]{RN38}
Aniek~F Markus, Jan~A Kors, and Peter~R Rijnbeek.
\newblock The role of explainability in creating trustworthy artificial
  intelligence for health care: a comprehensive survey of the terminology,
  design choices, and evaluation strategies.
\newblock \emph{Journal of Biomedical Informatics}, 113:\penalty0 103655, 2021.
\newblock ISSN 1532-0464.

\bibitem[Sundararajan et~al.(2017)Sundararajan, Taly, and Yan]{RN49}
Mukund Sundararajan, Ankur Taly, and Qiqi Yan.
\newblock Axiomatic attribution for deep networks.
\newblock In \emph{International conference on machine learning}, pages
  3319--3328. PMLR, 2017.
\newblock ISBN 2640-3498.

\bibitem[Hsu et~al.(2019)Hsu, Liu, and Tseng]{RN30}
En-Yu Hsu, Chien-Liang Liu, and Vincent~S Tseng.
\newblock Multivariate time series early classification with interpretability
  using deep learning and attention mechanism.
\newblock In \emph{Pacific-Asia Conference on Knowledge Discovery and Data
  Mining}, pages 541--553. Springer, 2019.

\end{thebibliography}

\end{document}